\documentclass[10pt,twocolumn,letterpaper]{article}

\usepackage[pagenumbers]{cvpr} 

\usepackage{subfiles}
\usepackage{multirow}
\usepackage{color}
\definecolor{cvprblue}{rgb}{0.21,0.49,0.74}
\usepackage[pagebackref,breaklinks,colorlinks,allcolors=cvprblue]{hyperref}

\newcommand{\FIG}[1]{\cref{#1}}
\newcommand{\TABLE}[1]{\cref{#1}}
\newcommand{\EQUA}[1]{\cref{#1}}
\newcommand{\SECTION}[1]{\cref{#1}}

\newcommand{\EG}{{\emph{e.g.}}}
\newcommand{\IE}{{\emph{i.e.}}}

\newcommand{\SOCIALLSTM}{Social-LSTM}
\newcommand{\SOCIALLSTMCITE}{\cite{alahi2016social}}
\newcommand{\SOCIALGAN}{S-GAN}
\newcommand{\SOCIALGANCITE}{\cite{gupta2018social}}

\newcommand{\TRAJECTRONPPCITE}{\cite{salzmann2020trajectron}}
\newcommand{\YNET}{$\textsf{Y}$-net}
\newcommand{\YNETCITE}{\cite{mangalam2020s}}

\newcommand{\STGCNN}{Social-STGCNN}
\newcommand{\STGCNNCITE}{\cite{mohamed2020social}}

\newcommand{\SPECTGNN}{SpecTGNN}
\newcommand{\SPECTGNNCITE}{\cite{cao2021spectral}}

\newcommand{\AGENTFORMER}{AF}
\newcommand{\AGENTFORMERCITE}{\cite{yuan2021agentformer}}

\newcommand{\MSN}{MSN}
\newcommand{\MSNCITE}{\cite{wong2021msn}}

\newcommand{\SEEM}{SEEM}
\newcommand{\SEEMCITE}{\cite{wang2022seem}}

\newcommand{\NSPSFM}{NSP-SFM}
\newcommand{\NSPSFMCITE}{\cite{yue2022human}}
\newcommand{\MUSEVAE}{MUSE-VAE}
\newcommand{\MUSEVAECITE}{\cite{lee2022muse}}

\newcommand{\EVMODEL}{E-V$^2$-Net}

\newcommand{\EVCITE}{\cite{wong2023another}}
\newcommand{\EQMOTION}{EqMotion}
\newcommand{\EQMOTIONCITE}{\cite{xu2023eqmotion}}

\newcommand{\FLOWCHAIN}{FlowChain}
\newcommand{\FLOWCHAINCITE}{\cite{maeda2023fast}}
\newcommand{\IMP}{IMP}
\newcommand{\IMPCITE}{\cite{shi2023representing}}
\newcommand{\SCMODEL}{SocialCircle}
\newcommand{\SCCITE}{\cite{wong2023socialcircle}}

\newcommand{\LGTRAJ}{LG-Traj}
\newcommand{\LGTRAJCITE}{\cite{chib2024lg}}
\newcommand{\RAN}{RAN}
\newcommand{\RANCITE}{\cite{dong2024recurrent}}

\newcommand{\UPDD}{UPDD}
\newcommand{\UPDDCITE}{\cite{liu2024uncertainty}}

\def\paperID{4828} 
\def\confName{CVPR}
\def\confYear{2025}

\title{Who Walks With You Matters: Perceiving Social Interactions with Groups for Pedestrian Trajectory Prediction}

\author{
    Ziqian Zou\quad
    Conghao Wong\quad
    Beihao Xia\quad
    Qinmu Peng\quad
    Xinge You\quad \\
    Huazhong University of Science and Technology\quad \\
    {\tt\footnotesize ziqianzoulive@icloud.com, conghaowong@icloud.com,} \\
    {\tt\footnotesize xbh\_hust@hust.edu.cn, pengqinmu@hust.edu.cn, youxg@mail.hust.edu.cn}
}

\begin{document}
\maketitle

\begin{abstract}

Understanding and anticipating human movement has become more critical and challenging in diverse applications such as autonomous driving and surveillance.
The complex interactions brought by different relations between agents are a crucial reason that poses challenges to this task.
Researchers have put much effort into designing a system using rule-based or data-based models to extract and validate the patterns between pedestrian trajectories and these interactions, which has not been adequately addressed yet.
Inspired by how humans perceive social interactions with different level of relations to themself, this work proposes the GrouP ConCeption (short for GPCC) model composed of the Group method, which categorizes nearby agents into either group members or non-group members based on a long-term distance kernel function, and the Conception module, which perceives both visual and acoustic information surrounding the target agent. 
Evaluated across multiple datasets, the GPCC model demonstrates significant improvements in trajectory prediction accuracy, validating its effectiveness in modeling both social and individual dynamics. 
The qualitative analysis also indicates that the GPCC framework successfully leverages grouping and perception cues human-like intuitively to validate the proposed model's explainability in pedestrian trajectory forecasting.
Codes are available at \url{https://github.com/livepoolq/groupconception}.
\end{abstract}

\section{Introduction}

\begin{figure}[t]
    \centering
    \includegraphics[width=1.0\linewidth]{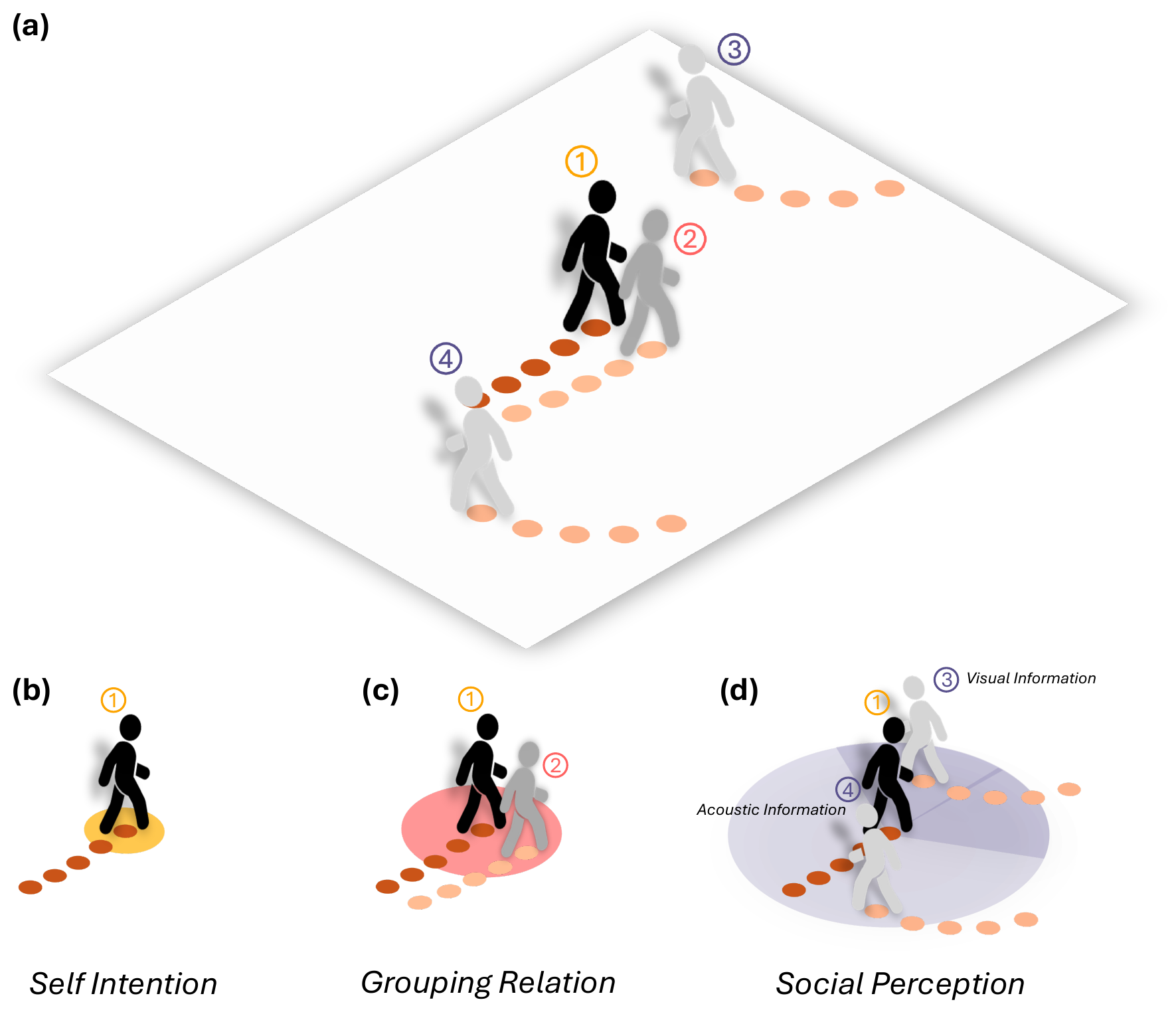}
    \caption{
        We divide factors that influence our decision to move and behave into three parts, as shown in (b), (c), and (d), according to scene (a).
 (b) \textbf{Self intention} denotes the agent's self-will of destination.
 (c) \textbf{Grouping relation} denotes other group members' impact on the target agent.
 (d) \textbf{Social perception} denotes how the target agent perceives surrounding agents with visual and acoustic information.
 The main contributions in this work originate from modeling these three factors and how they cooperate to improve the trajectory prediction performance.
    }
    \label{fig_motivation}
\end{figure}

Pedestrian trajectory prediction is a critical and challenging task in various applications, such as autonomous driving and surveillance, where understanding and anticipating human movement can significantly enhance safety and efficiency. 
Traditional trajectory prediction models often learn social interactions brought by different agents with the same paradigm, overlooking the influence of prior knowledge, such as grouping relations and interaction dynamics, on pedestrian movement.
In light of this limitation, this paper introduces the GPCC model (GrouP ConCeption), which incorporates three essential factors to improve trajectory prediction accuracy: self-intention, grouping relations, and social perception(\FIG{fig_motivation}).

Self-Intention indicates that each pedestrian has an intrinsic goal or intended destination, which determines their basic movement. 
The incorporation of this self-driven intent ensures that the model respects the primary individual motivation inherent in a pedestrian's trajectory, thereby providing a base prediction that is aligned with the person's objectives.
Grouping Relations are widely observed in real-world settings, \IE, pedestrians often move in groups (\EG, with family or friends).
Members within these groups exert influence over one another's movements. Recognizing and modeling these ``related'' group dynamics allows the model to predict the consistent movement patterns often observed in group scenarios more accurately. Revealing this relation presents a significant improvement over models that treat all neighboring agents with the same paradigm in effectiveness and explainability.
Social Perception represents that pedestrians continuously adjust their paths based on the presence and behavior of other ``unrelated'' pedestrians, \IE, agents not belonging to the same group. 
This factor incorporates visual and acoustic cues within and out of the pedestrian's field of view (short for FOV ) , enabling the model to capture the target agent's responses to other agents' movements, such as avoiding collisions or making path adjustments.

By integrating these three factors, the GPCC model offers a self-aware, group-aware, and socially-aware trajectory prediction approach, capable of accounting for both individual motivations and the different levels of the interactive context of each pedestrian's surrounding agents and allows the model to better predict trajectories by considering both "related" and "unrelated" agents in the environment.

In summary, we contribute (1) The Group method that divides neighboring agents into the grouping members and ones out of this group based on long-term distance kernel function; (2) The Conception module that perceives visual and acoustic information from different regions around the target agent based on vision range of human eyes; (3) The validation of the effectiveness and explainability of the proposed GPCC model by evaluating it on different datasets and analyzing how its components function to co-contribute to improve the model's trajectory prediction performance.

\section{Related Works}

~\\\noindent\textbf{Pedestrian Trajectory Prediction.}
Pedestrian trajectory prediction has evolved considerably, beginning with traditional rule-based models\cite{helbing1995social,pellegrini2009youll,luber2010people,barata2021sparse,mehran2009abnormal} and advancing to complex deep learning methods that aim at capturing the high-dimensional features of social interactions with the rise of deep learning.
These data-based methods\cite{pei2019human,gupta2018social,vemula2017modeling,kothari2020human} introduced recurrent neural networks and generative adversarial networks to better capture the serial dependencies in trajectory prediction, which improved trajectory accuracy by learning latent social behaviors from data.
Graph neural networks (GNNs) have also been adopted to model social relationships in trajectory prediction\cite{cao2020spectral,li2022intention,su2022trajectory,kim2024higher}, which allow for more expressive and context-aware social interaction modeling, as they capture the dynamic dependencies between individuals in the scene.

~\\\noindent\textbf{Holistic Social Interaction Modeling.} 
In most previous works\cite{pei2019human,gupta2018social,vemula2017modeling,wong2023socialcircle}, they adopt a unified paradigm to represent social interactions across all pedestrians in the scene using certain properties of deep learning models. 
These approaches typically apply a single, generalized representation for all agents, thereby treating social interactions as homogenous across the entire scene. 
Although this method is often equipped with considerable prediction performance, it often overlooks critical cues in social relationships, such as group dynamics or specific relational contexts that differ among individuals. 
This limitation can lead to less accurate predictions in complex social environments where group relationships play a significant role.

~\\\noindent\textbf{Multi-Level Social Interaction Modeling.} 
To address some of the information loss inherent in holistic representations, multi-level approaches attempt to model social interactions at different levels of granularity. 
These methods\cite{xu2022groupnet,bae2022learning,kim2024higher,wong2021msn} incorporate more specific relational information, such as distinguishing between various types of social connections, which partially alleviates the issue of missing social relationship details. 
However, the majority of these approaches still rely heavily on black-box models driven by data, offering limited interpretability, making it challenging to understand the model's internal decision-making processes.

The proposed GPCC model adopts the multi-level interaction modeling paradigm with an explainable feature-extracting process, improving the interpretability of the most current data-based methods and acquiring considerable trajectory prediction accuracy.

\section{Method}
\label{method}

\subsection{Problem Formulation}
Pedestrian trajectory prediction task can be described as forecasting trajectories in the future time range $T_{\mathrm{future}}$ based on what we observed in the past time range $T_{\mathrm{past}}$.
Let current time be the origin of the time axis. This work mainly concerns predicting the 2D coordinates of the target agent $i$ when $t=\{T, 2T, 3T, \dots, n_f T\}$ based on all $N-1$ neighboring agents' and agent $i$'s past positions when $t=\{-(n_p -1)T, -(n_p - 2)T, \dots, -T, 0\}$, where $T$ is the interval of sampling continuous trajectories into discrete sequences, $n_f T = T_{\mathrm{future}}$ and $(n_p - 1)T = T_{\mathrm{past}}$.
Formally, it aims at designing a network $\mathcal{N}$ that satisfies
$\hat{\mathbf{Y}}_i = \mathcal{N} \left( \mathbf{X}_i, \Omega_\mathrm{traj} \backslash \left\{\mathbf{X}_i\right\}, \mathbf{C}\right)$,
where $\mathbf{X}_i$ indicates the target agent $i$'s historical coordinate sequence, $\Omega_\mathrm{traj}$ is a \emph{set} that contains all $N$ agents' historical coordinate sequences, and $\mathbf{C}$ represents other information that can be used as input of the network $\mathcal{N}$ to predict agent $i$'s future sequence $\hat{\mathbf{Y}}_i$.

\subsection{Group Method Modeling}
It aligns with our instinct that pedestrians walking in a \textbf{group} which contains other ``related'' agents interact with other agents differently compared with those walking alone among other strangers.

~\\\noindent\textbf{Long-Term Distance Kernel Function.}
To better illustrate this grouping relation, we introduce the long-term distance kernel function denoted as $K (\cdot)$ to distinguish group members of the target agent from all neighboring agents by calculating the \emph{overall} distance sum between them during all observed or past time $T_{\mathrm{past}}$.

\begin{figure}[t]
    \centering
    \includegraphics[width=1.0\linewidth]{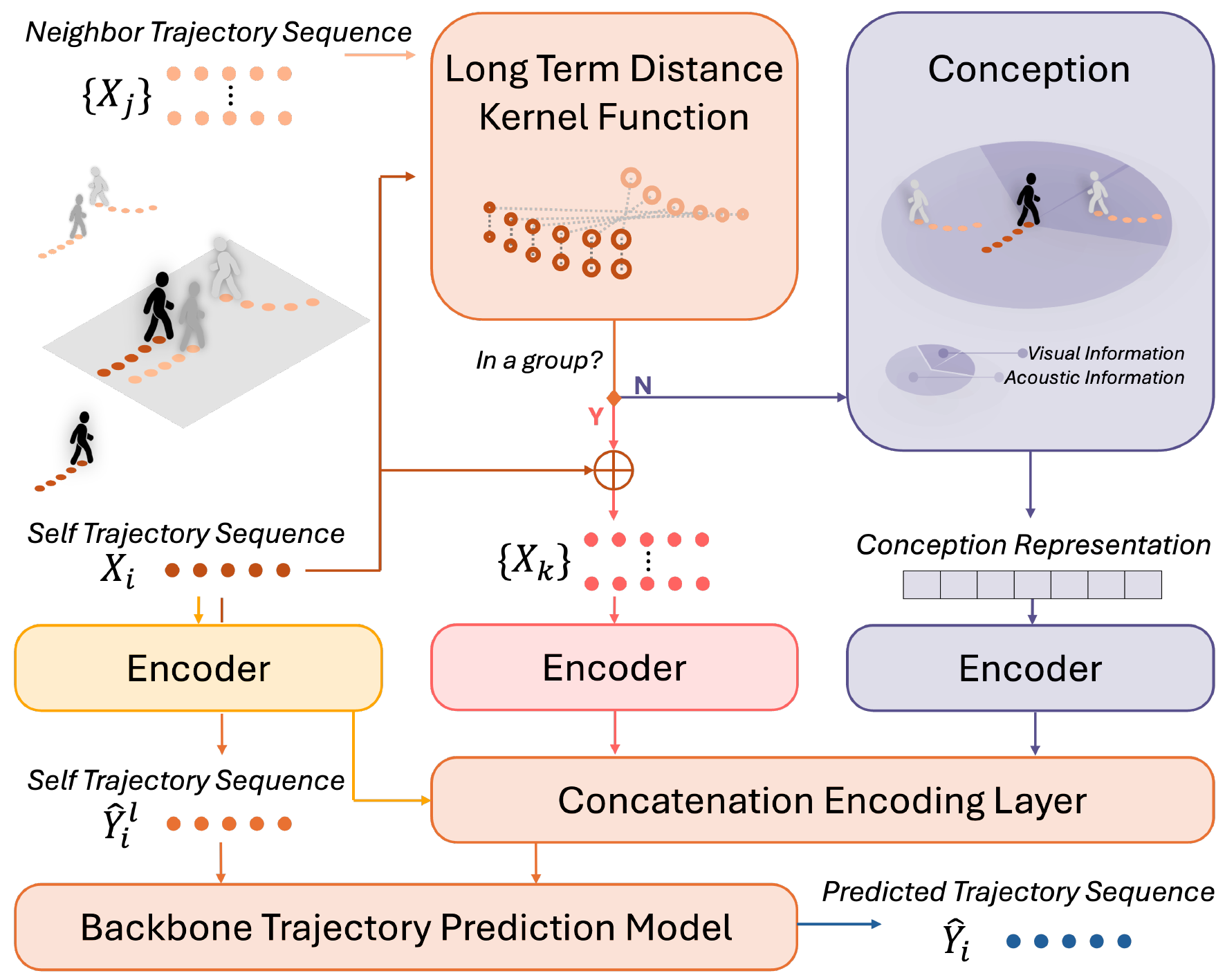}
    \caption{
        Schema of the modeling process of the proposed GPCC model and the computation pipeline of the Group method and the Conception module.
 The Group Networks divide the neighbors into two sets based on long-term distance kernel function $K$.
 A group encoder encodes one sequence set within the Group method, and the Conception module perceives the other to gain encoded Conception representation.
 The concatenation of these two features and the encoded self-trajectory sequence is the backbone trajectory prediction model's input feature to forecast the target agent's future trajectories.
    }
    \label{fig_method}
\end{figure}

Formally, the long-term distance kernel function $K (\cdot)$ applying to the target agent $i$ and a neighboring agent $j$ can be described as
\begin{equation}
    K (i, j) = 
    \begin{cases}
        1, ~~\sum_{t \in \Omega_{T_\mathrm{past}}} \Vert \mathbf{p}_j^t - & \mathbf{p}_i^t \Vert_2 \le d_\mathrm{m} \\
        0, & otherwise  
    \end{cases}, 
\end{equation}
where $\mathbf{p}_j^t$ and $\mathbf{p}_i^t$ represent the positions of the neighboring agent and the target agent at time $t$, which belongs to a set $\Omega_{t_\mathrm{past}}$ containing all $T$-sampled $n_p$ observed frames.
The manually designed $d_\mathrm{m}$ represents the threshold to distinguish group members from all agents \footnote{See discussions of designing the threshold $d_\mathrm{m}$ in \textcolor{blue}{Supplementary Material}.}.

Agent $k$ within the target agent $i$'s group can be described as the set $\mathbf{N}^i_{\mathrm{group}}$, which satisfies
\begin{equation}
    j \in \mathbf{N}^i_{\mathrm{group}}, \forall K(i,j)=1.
\end{equation}
We design a trajectory encoding network $g(\cdot)$ containing two fully connected layers with $d=32$ output units.
ReLU activation is used in the first layer, while tanh is used in the second.
\begin{equation}
    \label{eq_group}
    \mathbf{f}^i_{\mathrm{group}} = g
    \left( \mathbf{X}_k \right), \forall k \in \mathbf{N}^i_{\mathrm{group}}.
\end{equation}
As shown in \FIG{fig_method} and \EQUA{eq_group}, all agent $k$'s trajectories are concatenated with the target agent $i$'s trajectory and then embedded through trajectory encoding networks $g(\cdot)$.

\subsection{Conception Module Modeling}
The Conception module is introduced for target agents to perceive ``unrelated'' neighbors around them.

As shown in \FIG{fig_method}, pedestrians perceive other agents with broader information within the range of FOV.
They might intuitively observe whether the other pedestrians' motions and behavior will influence their decision to choose a routine. 
For agents out of the FOV, pedestrians might not be able to perceive such broad information, but their movements could still be noticed when approaching the target pedestrian.
To illustrate this difference, we use two modeling strategies for them.

~\\\noindent\textbf{Visual Information Strategy.}
Agent $j$ within the target pedestrian $i$'s FOV angle $\theta_\mathrm{FOV}^i$ can be described as the set $\mathbf{N}(\theta_\mathrm{FOV}^i)$, which satisfies
\begin{equation}
 \vert \theta^j - \theta^i_\mathrm{move} \vert \le \frac{\theta_\mathrm{FOV}^i}{2}, \forall j \in \mathbf{N}(\theta_\mathrm{FOV}^i).
\end{equation}
$\theta^j$ represents neighboring agent $j$'s position angle relative to the target agent $i$'s position, which can be seen as the original point.
Agent $i$'s moving direction $\theta^i_\mathrm{move}$ is approximated as position vector from $t_{-1}=-T$ to $t_0=0$:
\begin{equation}
    \theta^i_\mathrm{move} = \mathrm{atan2} \left(\mathbf{p}_i^{t_0} - \mathbf{p}_i^{t_{-1}}\right).
\end{equation}
For all agent $j \in \mathbf{N}(\theta_\mathrm{FOV}^i)$, we mainly consider their distance, relative moving direction, and velocity to the target agent $i$.
The FOV angle $\theta_\mathrm{FOV}^i$ is equally divided into two partitions representing the left and right side of the target agent, and the division line is defined as the moving direction of him.
We calculate the average of these three visual information acquired in each partition of FOV regions.
The computation process of left and right regions ($\theta^i_\mathrm{left}$ and $\theta^i_\mathrm{right}$) are symmetrical, and we represent the left region here as an example. We define a set $\mathbf{N}(\theta_\mathrm{left}^i)$ composed of ``unrelated'' agents in the left region and a function $n(\cdot)$ to calculate number of agents in a set.

We use the average Euclidean distance of the neighboring agents and the target agent at $t_0=0$ moment to model distance information, which is calculated by
\begin{equation}
 {dis} (\theta_\mathrm{left}^i)= 
    \sum_{j \in \mathbf{N}(\theta_\mathrm{left}^i)} {\left\Vert \mathbf{p}_j^{t_0} - \mathbf{p}_i^{t_0} \right\Vert _2}/{n\left(\mathbf{N}(\theta_\mathrm{left}^i)\right)}.
\end{equation}

Similarly to the distance factor, the average relative moving direction factor is calculated by
\begin{equation}
 {dir} (\theta_\mathrm{left}^i)= 
    \sum_{j \in \mathbf{N}(\theta_\mathrm{left}^i)} {\left\vert \theta^j_\mathrm{move} - \theta^i_\mathrm{move} \right\vert}/{n\left(\mathbf{N}(\theta_\mathrm{left}^i)\right)}.
\end{equation}

The Conception module takes the average velocity of all agents within the FOV to model the velocity factor:
\begin{equation}
 {vel} (\theta_\mathrm{left}^i)= 
    \sum_{j \in \mathbf{N}(\theta_\mathrm{left}^i)} {\left\Vert \mathbf{p}_j^{t_{-(n_p - 1)}} - \mathbf{p}_j^{t_0} \right\Vert _2}/{n\left(\mathbf{N}(\theta_\mathrm{left}^i)\right)}.
\end{equation}

~\\\noindent\textbf{Acoustic Information Strategy.}
For $i$'s neighboring agents who are out of the FOV, the Conception module only considers their distance information, which can be acquired visually and acoustically:
\begin{equation}
 {dis} (\theta_\mathrm{rear}^i)= 
    \sum_{j \in \mathbf{N}(\theta_\mathrm{rear}^i)} {\left\Vert \mathbf{p}_j^{t_0} - \mathbf{p}_i^{t_0} \right\Vert _2}/{n\left(\mathbf{N}(\theta_\mathrm{rear}^i)\right)}.
\end{equation}
The three partitions, \IE \emph{left}, \emph{right}, and \emph{rear}, act as position encoding of neighboring agents, allowing the module to perceive their position information from an intuitive perspective rather than simply pooling features around the target agent.
Notably, all average calculations proceed in each partition, and all factors are concatenated in the order of right(short for r in \EQUA{eq_concat}), left and rear as below:
\begin{equation}
\begin{aligned}
    \label{eq_concat}
 \mathbf{r}_{\mathrm{con}}^i =\left(  
 {dis} (\theta_\mathrm{r}^i),
 {dir} (\theta_\mathrm{r}^i), 
 {vel} (\theta_\mathrm{r}^i), 
 ...,
 {dis} (\theta_\mathrm{rear}^i)
 \right)^T.
\end{aligned}
\end{equation}
The Conception representation $\mathbf{r}_{\mathrm{con}}^i$ is embedded into high-dimensional feature $\mathbf{f}^i_{\rm{con}} \in \mathbb{R}^{n_p \times 2d}$ by emdedding layer $h(\cdot)$, \IE, $\mathbf{f}^i_{\rm{con}} = h\left(\mathbf{r}_{\mathrm{con}}^i\right)$.

\subsection{Fusion Strategy and Overall Modeling}
In our work, agent $i$'s observed trajectory $\mathbf{X}_i$ is also first embedded into high-dimensional feature denoted as $\mathbf{f}^i_{\rm{self}} \in \mathbb{R}^{n_p \times d}$ by emdedding layer $f(\cdot)$, \IE, $\mathbf{f}^i_{\rm{self}} = f\left(\mathbf{X}_i\right)$.

As shown in \FIG{fig_method}, the GrouP ConCeption model (short for GPCC) we propose takes the fused vector $\mathbf{f}^i$ containing Conception feature $\mathbf{f}^i_{\rm{con}}$ (representation of social interactions with ``unrelated'' neighboring agents), self-centered feature $\mathbf{f}^i_{\rm{self}}$ (representation of ego trajectory) and Group feature $\mathbf{f}^i_{\mathrm{group}}$ (representation of group members' trajectories) gained in \EQUA{eq_group} as input to learn to predict future trajectories. 
See \FIG{fig_motivation,fig_method} for a more vivid illustration of this combination.
The $\mathbf{f}^i$ is fused by
\begin{equation}
    \label{eq_fuse}
 \mathbf{f}^i = \tanh \left(
 \mathbf{W}_{\rm{fuse}}~{\rm{Concat}} \left(
 \mathbf{f}^i_{\rm{con}}, 
 \mathbf{f}^i_{\rm{self}}, \mathbf{f}^i_{\mathrm{group}}
 \right)
 \right).
\end{equation}
$\mathbf{W}_{\rm{fuse}}$ are the trainable weights in \EQUA{eq_fuse} and trainable bias is omitted in equation for brevity.

We take Transformer\cite{vaswani2017attention} and multi-style trajectory generation module in MSN\cite{wong2021msn} as backbone trajectory prediction model.
As depicted in \FIG{fig_method}, the Transformer encoder learns the high-dimensional information of $\mathbf{f}^i$ by computing self-attention over $n_p$ steps (number of observation frames). The decoder takes the encoder's output features as keys and queries and the linear fitting trajectory $\hat{\mathbf{Y}}_i^l$ of the target agent as values.
The Transformer outputs a feature for the trajectory generation module\cite{wong2021msn} to generate trajectories.
The pipeline is 
\begin{equation}
 \hat{\mathbf{Y}}_i = B_{\rm{prediction}} \left(
 \mathbf{f}^i, \hat{\mathbf{Y}}_i^l
 \right).
\end{equation}

~\\\noindent\textbf{Training.}
The proposed GPCC model is trained with  $\ell_2$ loss and does not introduce any new loss function, which is calculated as
\begin{equation}
    \ell_2 \left(
 \mathbf{Y}^i,
 \left\{\hat{\mathbf{Y}}^i_k\right\}
 \right) = \min \left\{ \left\Vert
 \mathbf{Y}^i - \hat{\mathbf{Y}}^i_k
 \right\Vert_2 \vert 1\le k \le K \right\}
\end{equation}
under the \emph{best-of-$K$} \cite{gupta2018social} validation.

\section{Experiments}

\subsection{Experimental Settings}

\noindent\textbf{Datasets.}
To evaluate the trajectory prediction performance of the proposed GPCC model, we utilize four datasets: (1) \textbf{ETH-UCY}~\cite{pellegrini2009youll,lerner2007crowds}, (2) \textbf{Stanford Drone Dataset} (short for SDD)~\cite{robicquet2016learning}, (3) \textbf{nuScenes}\footnote{See comparisons of the nuScenes dataset in \textcolor{blue}{Supplementary Material}. }~\cite{caesar2020nuscenes}, and (4) \textbf{NBA SportVU} (short for \textbf{NBA}\footnote{See comparisons of the NBA dataset in \textcolor{blue}{Supplementary Material}. })~\cite{linou2016nba}.

(1) \textbf{ETH-UCY}~includes several videos captured in pedestrian walking scenes.
We follow the \emph{leave-one-out} \cite{alahi2016social} strategy to train and validate models with $\left\{n_p = 8, n_f = 12\right\}$ and a sampling interval of $T = 0.4$s.

(2) \textbf{SDD}~comprises 60 video recordings obtained from an aerial perspective of the campus.
The agents, which fall into different categories (e.g., pedestrian, bicyclist, skateboarder, cart, car, and bus), have been labeled in pixels.
The 60\% of videos were designated for training, the 20\% for validation, and the 20\% for testing, with the same settings $\{n_p, n_f, T\} = \{8, 12, 0.4\mathrm{s}\}$ as \cite{liang2020simaug}.

~\\\noindent\textbf{Metrics.}
We use the same \emph{best-of-K} strategy as~\cite{alahi2016social,gupta2018social} to evaluate the quantitative performance of the proposed model with the metrics of best Average/Final Displacement Error over $K$ generated trajectories,\IE, $minADE_K$ and $minFDE_K$.
They are calculated as follows:
\begin{align}
 minADE_K(i)
    &= \min_{1 \leq k \leq K}  \frac{1}{n_f}
 \left\{ \sum_{t = T}^{n_f T}
 \left\Vert
 \mathbf{p}^t_i - 
 \hat{\mathbf{p}^t_i}(k)
 \right\Vert_2 \right\}, \\
 minFDE_K(i)
    &= \min_{1 \leq k \leq K, t = n_f T} \left\{
 \left\Vert
 \mathbf{p}^{t}_i - 
 \hat{\mathbf{p}^{t}_i}(k)
 \right\Vert_2 \right\}.
\end{align}

~\\\noindent\textbf{Implementation Details.}
GPCC model and its variations are trained on one NVIDIA GeForce RTX 3090. 
The FOV angle of the Conception module is set to be $180^{\circ}$(discussion of choosing the FOV angle for the Conception module in \SECTION{subsec_discussions}).
Following \cite{zhang2020social}, trajectories are preprocessed by moving to (0, 0).
We use the Adam optimizer with a learning rate 0.0002 to train our models, and the batch size is 1000 for 200 epochs.

\subsection{Comparisons}

We compare the proposed GPCC model with several state-of-the-art methods as shown in \TABLE{tab_ethucy_comparisons,tab_sdd_comparisons}.
The introductions of grouping relations are also included in some of these methods\cite{bae2022learning,xu2022groupnet}.

\begin{table*}[htbp]
    \centering
    \footnotesize
    \begin{tabular}{|l|cccccc|}
        \toprule
        Model & eth & hotel & univ & zara1 & zara2 & Average \\
    
        \midrule
        GP-Graph-STGCNN\cite{bae2022learning}~(2022) & 0.48/0.77 & 0.24/0.40 & 0.29/0.47 & 0.24/0.42 & 0.23/0.40 & 0.29/0.49 \\
        GP-Graph-PECNet\cite{bae2022learning}~(2022) & 0.56/0.82 & 0.18/0.26 & 0.31/0.46 & 0.23/0.40 & 0.17/0.27 & 0.29/0.44 \\
        GroupNet+CVAE\cite{xu2022groupnet}~(2022) & 0.46/0.73 & 0.15/0.25 & 0.26/0.49 & 0.21/0.39 & 0.17/0.33 & 0.25/0.44 \\
        GroupNet+Trajectron++\cite{xu2022groupnet}~(2022) & 0.38/0.74 & \underline{0.11}/0.20 & \underline{0.19}/\underline{0.40} & \textbf{0.14}/0.32 & \textbf{0.11}/0.25 & 0.19/0.38 \\
        \YNET\YNETCITE~(2021) & 0.28/\textbf{0.33} & \textbf{0.10}/\textbf{0.14} & 0.24/0.41 & 0.17/0.27 & \underline{0.13}/\textbf{0.22} & \underline{0.18}/\textbf{0.27} \\
        \SEEM\SEEMCITE~(2023) & 0.62/1.20 & 0.61/1.21 & 0.50/1.04 & 0.31/0.61 & 0.36/0.68 & 0.48/0.95 \\
        \EQMOTION\EQMOTIONCITE~(2023) & 0.40/0.61 & 0.12/0.18 & 0.23/0.43 & 0.18/0.32 & 0.13/0.23 & 0.21/0.35 \\
        \MSN\MSNCITE~(2023) & 0.27/0.41 & \underline{0.11}/0.17 & 0.28/0.48 & 0.22/0.36 & 0.18/0.29 & 0.21/0.34 \\
        LMTraj-SUP\cite{bae2024can}~(2024) & 0.65/1.04 & 0.26/0.46 & 0.57/1.16 & 0.51/1.01 & 0.38/0.74 & 0.48/0.88 \\
        EigenTrajectory+HighGraph\cite{kim2024higher}~(2024) & 0.33/0.56 & 0.13/0.21 & 0.23/0.47 & 0.19/0.33 & 0.15/0.25 & 0.21/0.36 \\
        \LGTRAJ\LGTRAJCITE~(2024) & 0.38/0.56 & \underline{0.11}/0.17 & 0.23/0.42 & 0.18/0.33 & 0.14/0.25 & 0.20/0.34 \\
        \SCMODEL\SCCITE~(2024) & \underline{0.25}/\underline{0.38} & 0.12/\textbf{0.14} & 0.23/0.42 & 0.18/0.29 & \underline{0.13}/\textbf{0.22} & \underline{0.18}/\underline{0.29} \\
        \UPDD\UPDDCITE~(2024) & \textbf{0.22}/0.42 & 0.17/0.30 & \textbf{0.14}/\textbf{0.28} & \underline{0.16}/0.30 & 0.14/0.31 & \textbf{0.17}/0.32 \\
        
        \midrule
        GPCC~\textbf{(Ours)} & \underline{0.25}/\underline{0.38} & \textbf{0.10}/\underline{0.15} & 0.25/0.44 & 0.17/0.28 & \underline{0.13}/\textbf{0.22} & \underline{0.18}/\underline{0.29} \\
    
        \bottomrule
    \end{tabular}
    \caption{
        Comparisons to other state-of-the-art methods on ETH-UCY under $\{n_p, n_f, T\} = \{8, 12, 0.4\mathrm{s}\}$.
 Metrics are ``ADE/FDE'' (\emph{best-of-20}) in meters, and lower ADE and FDE represent better performance.
 \textbf{Bold}:Best, \underline{Underline}:Second Best.
    }
    \label{tab_ethucy_comparisons}
\end{table*}

\begin{table}[htbp]
    \centering
    \footnotesize
    \begin{tabular}{|lc|}
        \toprule
        Model & SDD \\

        \midrule
        \SPECTGNN\SPECTGNNCITE~(2021) & 8.21/12.41 \\
        \YNET\YNETCITE~(2021) & 7.85/11.85 \\
        GP-Graph-STGCNN\cite{bae2022learning}~(2022) & 10.6/20.5 \\
        GP-Graph-PECNet\cite{bae2022learning}~(2022) & 9.1/13.8 \\
        GroupNet+PECNet\cite{xu2022groupnet}~(2022) & 9.65/15.34 \\
        GroupNet+CVAE\cite{xu2022groupnet}~(2022) & 9.31/16.11 \\
        \NSPSFM\NSPSFMCITE~(2022) & 6.52/10.61 \\
        \MUSEVAE\MUSEVAECITE~(2022) & \textbf{6.36}/11.10 \\
        \FLOWCHAIN\FLOWCHAINCITE~(2023) & 9.93/17.17 \\
        \IMP\IMPCITE~(2023) & 8.98/15.54 \\
        LMTraj-SUP\cite{bae2024can}~(2024) & 17.5/34.5 \\
        \RAN\RANCITE~(2024) & 10.97/19.95 \\
        \LGTRAJ\LGTRAJCITE~(2024) & 7.80/12.79 \\
        \UPDD\UPDDCITE~(2024) & 6.59/13.90 \\
        \SCMODEL\SCCITE~(2024) & 6.54/\underline{10.36} \\
        
        \midrule
        GPCC~\textbf{(Ours)} & \underline{6.39}/\textbf{10.17} \\

        \bottomrule
    \end{tabular}
    \caption{
        Comparisons to other state-of-the-art methods on SDD under $\{n_p, n_f, T\} = \{8, 12, 0.4\mathrm{s}\}$.
 Notably, metrics are ``ADE/FDE'' (\emph{best-of-20}) in pixels rather than meters compared to ETH-UCY, and lower values indicate better prediction performance.
    }
    \label{tab_sdd_comparisons}
\end{table}

(1)\textbf{ETH-UCY.}
ETH-UCY contains pedestrian trajectories only, and we can observe that the proposed GPCC model has competitive performance compared with other state-of-the-art methods.
As shown in \TABLE{tab_ethucy_comparisons}, the GPCC model has competitive performance compared with other state-of-the-art methods.
Notably, it outperforms the outstanding UPDD by 9.4\% FDE.
Compared with other grouping-related methods\cite{bae2022learning,xu2022groupnet}, the GPCC model also presents considerable performance.
Overall, the performance of GPCC has been validated on ETH-UCY.

(2)\textbf{SDD.}
SDD contains diverse scenes where more types of agents move and behave compared with ETH-UCY.
As shown in \TABLE{tab_sdd_comparisons}, the proposed GPCC model reaches the best prediction accuracy in gaining better FDE by 2.1\% lower than the second best method SocialCircle\cite{wong2023socialcircle} and the second best ADE by only 0.4\% worse than the best-ADE method\MUSEVAECITE.
The results on SDD further verify the GPCC model's capability of handling more complex scenarios.

\subsection{Ablation Study}
\label{ablation_study}

We conduct ablation experiments\footnote{See results of SDD, NBA, and nuScenes in \textcolor{blue}{Supplementary Material}.} to further validate the effectiveness of different modules in our GPCC model as shown in \TABLE{tab_ethucy_variations}.

Comparing v3 (disabling both Group method and Conception module) and v0 (original GPCC model) from \TABLE{tab_ethucy_variations}, we can see that disabling Group and Conception could lead to a significant prediction performance drop of up to 58.6\% larger FDE.
Interestingly, it reaches the most performance drop of all variations at v2, which disables the Group method only and utilizes the Conception module.
Thought of this phenomenon is that when applying the Conception module of the GPCC model without the Group method would lead the model to focus on the ``unrelated'' agents only, which could cause more significant errors compared to the strategy of ignoring both ``related'' and ``unrelated'' agents.

Therefore, these results\footnote{Including results in \textcolor{blue}{Supplementary Material}.} demonstrate the quantitative improvements in prediction performance brought by the Group method and the Conception module and illustrate how their functions vary with different datasets and variations.

\begin{table*}[htbp]
    \centering
    \footnotesize
    \begin{tabular}{|c|cc|ccccccc|cccccccccc}
        \toprule
        ID & Group & Conception &
        eth &
        hotel &
        univ &
        zara1 &
        zara2 &
        ETH-UCY &
        $\Delta$ETH-UCY \\

        \midrule
        v1 & \multicolumn{2}{|l|}{~~~~~$\bullet$~~~~~~~~~~~~~~~~~$\circ$} 
        & 0.49/1.02 & 0.11/0.17 & 0.30/0.54 & 0.18/0.30 & 0.14/0.23 & 0.24/0.45 & 33.3\%/55.2\%\\
        v2 & \multicolumn{2}{|l|}{~~~~~$\circ$~~~~~~~~~~~~~~~~~$\bullet$} 
        & 0.51/1.05 & 0.11/0.16 & 0.30/0.57 & 0.18/0.30 & 0.14/0.23 & 0.25/0.46 & 38.9\%/58.6\%\\
        v3 & \multicolumn{2}{|l|}{~~~~~$\circ$~~~~~~~~~~~~~~~~~$\circ$} 
        & 0.53/1.13 & 0.11/0.16 & 0.29/0.53 & 0.13/0.23 & 0.14/0.23 & 0.24/0.46 & 33.3\%/58.6\%\\

        \midrule
        v0 & \multicolumn{2}{|l|}{~~~~~$\bullet$~~~~~~~~~~~~~~~~~$\bullet$} 
        & 0.25/0.38 & 0.10/0.15 & 0.25/0.44 & 0.17/0.28 & 0.13/0.22 & 0.18/0.29 & 0.0\%/0.0\%\\

        \bottomrule
    \end{tabular}
    \caption{
      Ablation studies on ETH-UCY.
      ``$\circ$'' means disabling the corresponding method or module and ``$\bullet$'' indicates the opposite.
      ``ID'' represents a different variation index of the proposed GPCC model, and ``$\Delta$ETH-UCY'' indicates the percentage of \textbf{performance drops} compared to the full GPCC model(v0).
    }
    \label{tab_ethucy_variations}
  \end{table*}

\subsection{Time Efficiency}
\label{time_efficiency}

The parameter amount of the overall GPCC model is 2057950.
The average inference time is 277 milliseconds with batchsize 1000 and 26 milliseconds with batchsize 100 by running model on one Apple Mac Studio (M2 Max) on the ETH-UCY dataset (49ms with batchsize 1000 and 24ms with batchsize 100 on the SDD dataset.). 
Comparisons with other methods are demonstrated in \textcolor{blue}{Supplementary Material}.

\subsection{Qualitative Analysis}
\label{subsec_discussions}

The previous \SECTION{ablation_study,time_efficiency} present the quantitative efficiency of the proposed GPCC model. 
In this section, we present the qualitative results of GPCC, which numerical metrics might not reveal.

~\\\noindent\textbf{Analysis of Group Method and kernel Function $K$.}
We first analyze whether the Group method in our GPCC model works and visualize how the long-term distance kernel function enables the model to learn group relations between pedestrians through a family group scene in ETH-UCY zara1.

\begin{figure}[tbph]
    \centering
    \includegraphics[width=1.0\linewidth]{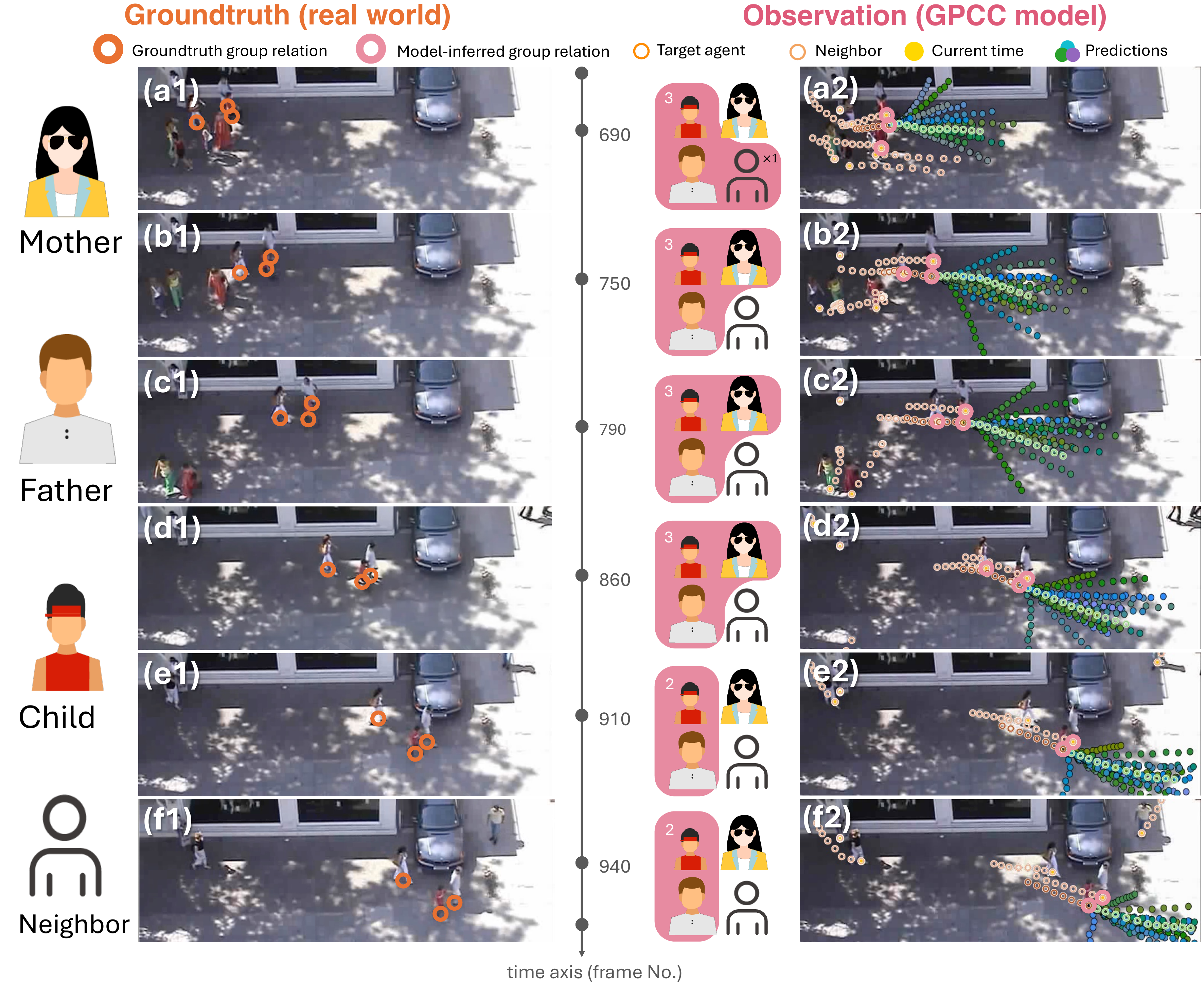}
    \caption{
        Illustration of how the Group method and the long-term distance kernel function $K$ work by comparing groundtruth with the GPCC model's observation.
 Figure legends on the left indicate a group containing three members ( Mother, Father, and Child) and ``unrelated'' neighbors in ETH-UCY ZARA1 scenes.
 We observe the Child as the target agent here.
 Indexes on the time axis are the frame number of the corresponding scene.
    }
    \label{fig_group_method}
\end{figure}

In the groundtruth(real world) scenario shown in \FIG{fig_group_method} (a1) - (f1), Mother, Father, and Child walk as a distinct group, which means they are ``related''.
Observing the zara1 clip from 28s to 43s can validate this grouping relation, which aligns with our manual-labeled relations between Mother, Father, and Child through orange concentric rings (short for the family group).
In the observation(GPCC model) scenario on the right side of \emph{frame axix} shown in \FIG{fig_group_method} (a2) - (f2), group relations derived from the Group method of GPCC model are marked with pink concentric rings.
We found some intriguing phenomena by comparing these two scenarios.

More specifically, the GPCC model does not consider the family group as a ``group'', and it divides Child, Father, and a pedestrian in a long red dress into a group of three members instead (\FIG{fig_group_method} (a1) and (a2)).
However, the GPCC model does a correct division of this family group in \FIG{fig_group_method} (b1) and (b2), (c1) and (c2), (d1) and (d2).
When Mother seems to be walking slower(\FIG{fig_group_method} (e1) and (f1)) than before, the model ``excludes'' her from the family group and only treats Father and Child as a group of two.

These splits conducted by the model align with human judgment to some extent because what the model receives are merely 2D coordinates during the past $n_p$ steps without semantic information of social relations between each agent.
Accordingly, the Group method could proceed with group divisions based on outputs of the kernel function $K$.
From the visualized results of this in \FIG{fig_group_method}, we could observe that the judgment made in this manner reflects the position relation between the target agent $i$ and another other agent $j$ in the observation time.
Since the prediction is forecasted with the information presented in the observation steps rather than in a much more comprehensive time range, this phenomenon of varying grouping splits with different observation time windows is in line with human instincts, which validates the effectiveness and the explainability of the Group method within the proposed GPCC model.

\begin{figure}[t]
    \centering
    \includegraphics[width=1.0\linewidth]{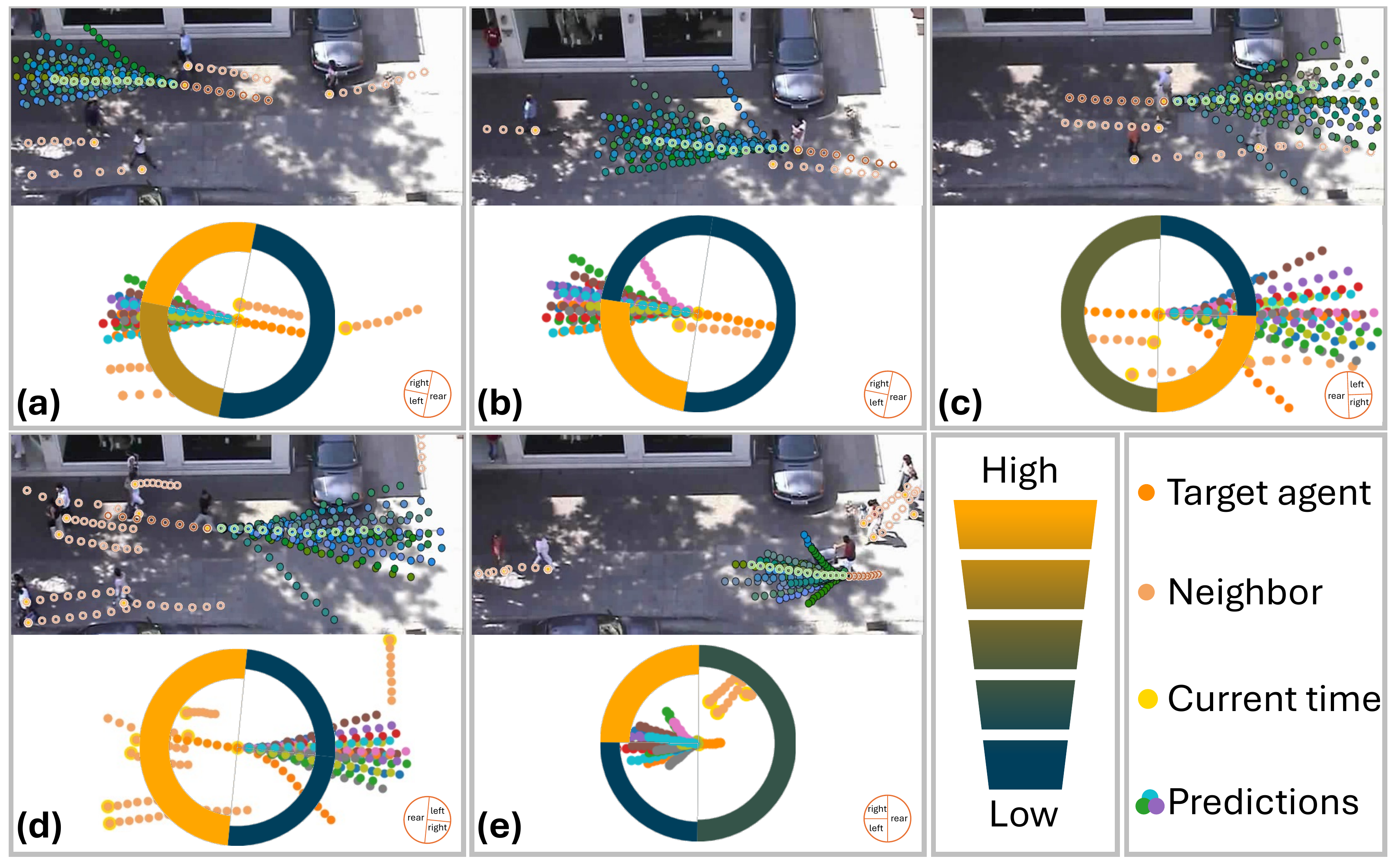}
    \caption{
        Visualization of attention value varies with different scenes.
 The model pays more attention to wider regions, whose color tends to be yellow accordingly.
    }
    \label{fig_conception_fov}
\end{figure}

~\\\noindent\textbf{Analysis of Conception Module and FOV Partitions\footnote{See detailed analysis of FOV partitions in \textcolor{blue}{Supplementary Material}.}}
We visualize these attention fan charts in different scenes in \FIG{fig_conception_fov}.
The target agent pays more attention on his left side when an agent approaches in the left partition(\FIG{fig_conception_fov} (b)), and the same goes in the right partition in \FIG{fig_conception_fov} (c).
When there is no agent in the front partition (combination of right and left partition) that could attract the target agent's eyes enough, he pays more attention to his rear partition, where there might be relatively more agents to raise his interest.

Results in \textcolor{blue}{Supplementary Material} validate the strategy of choosing the FOV angle in the Conception module and further support this analysis through visualizing the attention value fan charts.
Moreover, \FIG{fig_conception_fov} represents the effectiveness of the Conception module from an intuitive view.

~\\\noindent\textbf{Analysis of GPCC modeling.}
In \EQUA{eq_fuse}, we mentioned that the fused feature $\mathbf{f}^i$ to be sent to Transformer is composed of $\mathbf{f}^i_{\mathrm{self}}$, $\mathbf{f}^i_{\mathrm{group}}$ and $\mathbf{f}^i_{\mathrm{con}}$.
After analyzing the effectiveness and the explainability of the Group method and the Conception module, we mainly focus on the overall modeling of the proposed GPCC model in this part.
To sum up, the Group method enables the GPCC model to reveal grouping relations between agents, and the Conception module enables the model to perceive social interactions with those who do not belong to a group with the target agent.
They complement each other in contributing to the trajectory prediction performance of the GPCC model since the model is designed to consider relations within group members without neglecting the social interactions brought by the ``unrelated'' ones.
In \SECTION{ablation_study}, we conducted ablation experiments to verify the contributions of the Group method and the Conception module.
To further illustrate how these two strategies (Group and Conception) cooperate and ``interact'' with each other, we determine the contributions ratio function $r(\cdot)$ of the three features ($\mathbf{f}^i_{\mathrm{self}}$, $\mathbf{f}^i_{\mathrm{group}}$ and $\mathbf{f}^i_{\mathrm{con}}$) by calculating their ``energy'' flowing in the linear layer in \EQUA{eq_fuse}.
The pipeline includes:
\begin{equation}
 r(\cdot) = \frac{ \Vert
 \mathbf{m}(\cdot) * \cdot \Vert_2
 }{
            \sum_{\cdot \in \Omega_{\mathbf{f}}} 
 \Vert \mathbf{m}(\cdot) * \cdot \Vert_2
 }.
\end{equation}
$\Omega_{\mathbf{f}}$ denotes a set containing all three features, which satisfies $r(\mathbf{f}^i_{\mathrm{self}}) + r(\mathbf{f}^i_{\mathrm{group}}) + r(\mathbf{f}^i_{\mathrm{con}}) = 1 $.

\begin{figure}[t]
    \centering
    \includegraphics[width=1.0\linewidth]{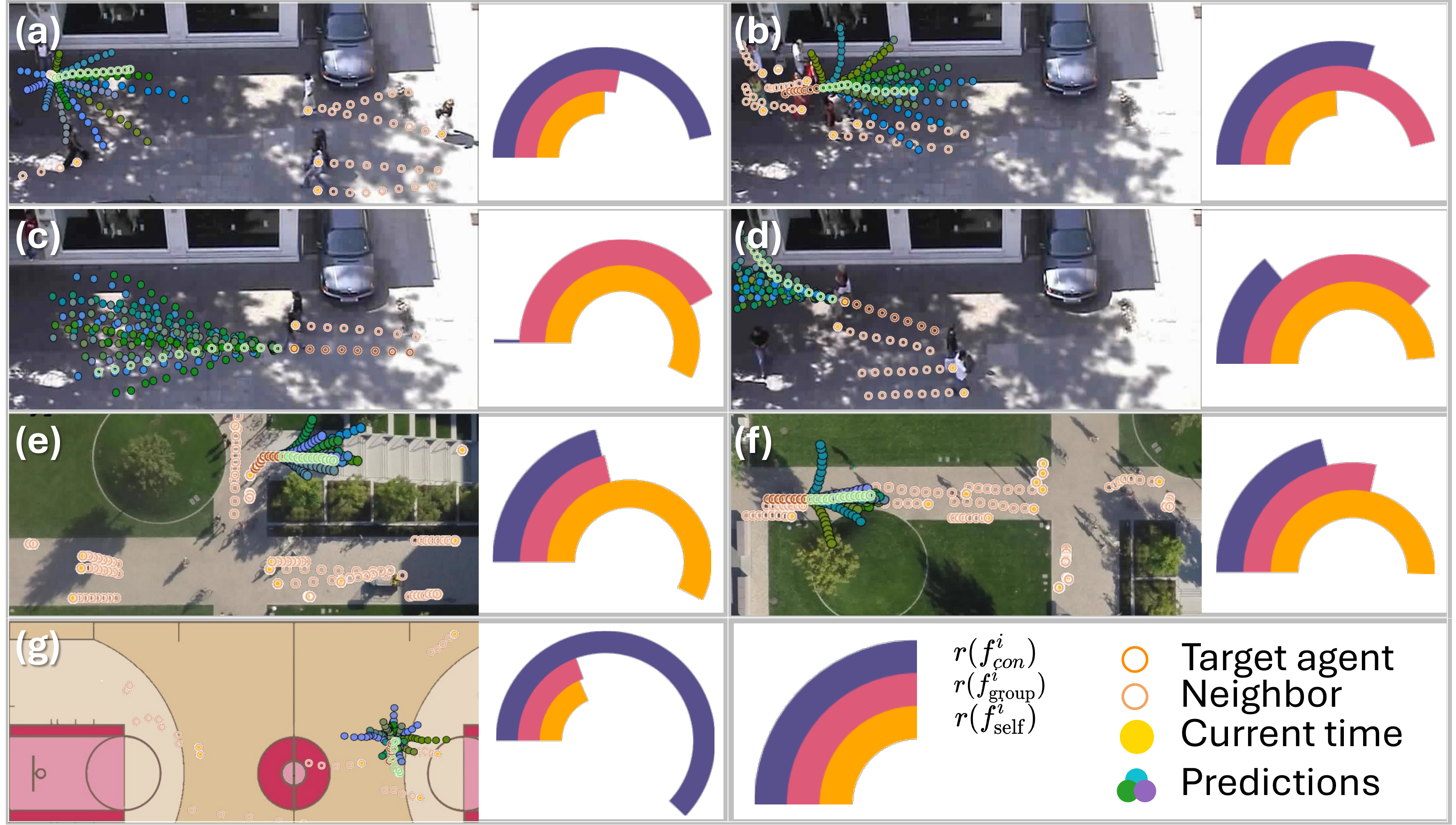}
    \caption{
        Visualized results of contribution ratio $r(\mathbf{f}^i_{\mathrm{self}})$, $r(\mathbf{f}^i_{\mathrm{group}})$ and $r(\mathbf{f}^i_{\mathrm{con}})$ in concentric fan chart form.
        The fan which has enormous angle value means more contribution and $r$ value of the \textcolor[RGB]{88,80,141}{\textbf{Conception}} feature, the \textcolor[RGB]{222,90,79}{\textbf{Group}} feature and the \textcolor[RGB]{255,166,00}{\textbf{Self}} feature are marked with the corresponding color. 
    }
    \label{fig_contributions}
\end{figure}

\begin{figure*}[tbph]
    \centering
    \includegraphics[width=1.0\linewidth]{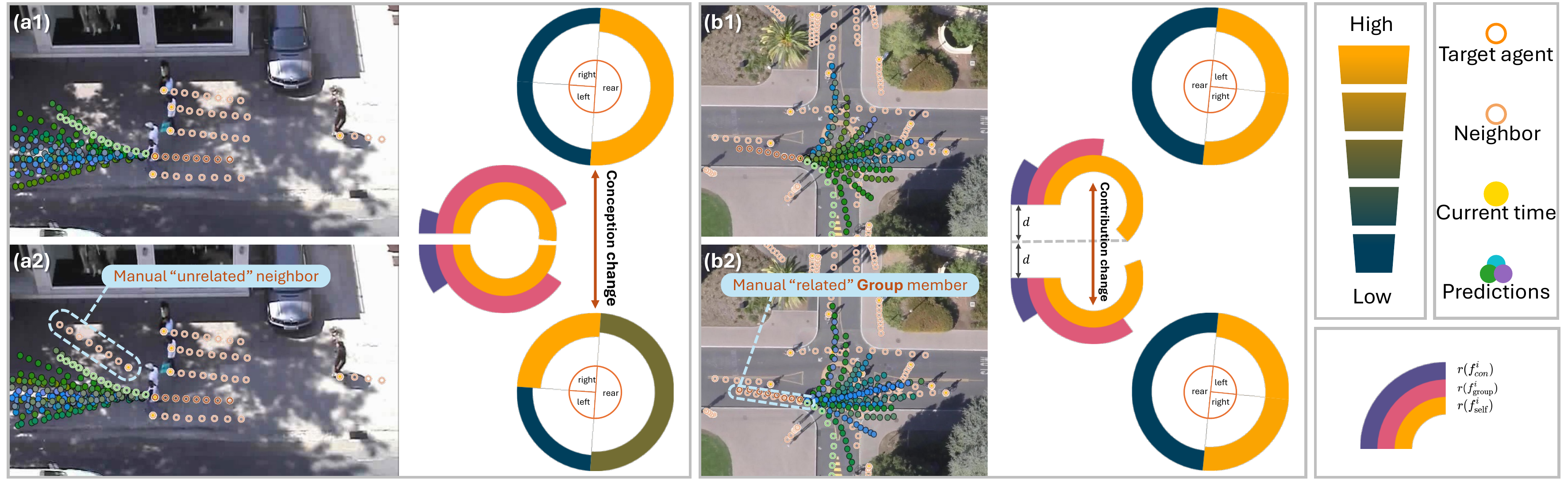}
    \caption{
        Visualization of the attention value change and the contribution change after placing different kinds of manual neighbors in ETH-UCY zara1 and SDD little0 scenes.
        Fan charts' settings are the same as \FIG{fig_conception_fov,fig_contributions}.
    }
    \label{fig_manual_neighbor}
\end{figure*}

In \FIG{fig_contributions}, we visualize the contribution of each feature on different datasets with concentric fan charts to be analyzed.
The order of the concentric fan is designed to be aligned with the relative positional relations in reality that $r(\mathbf{f}^i_{\mathrm{self}})$ is positioned at inner side and then comes the $r(\mathbf{f}^i_{\mathrm{group}})$ while the $r(\mathbf{f}^i_{\mathrm{con}})$ surrounds them.
It should be noted that the ratios are presented relevant only to the angle of each fan chart, and the figure legends in \FIG{fig_contributions} demonstrate an example of the equal contribution ratio of the three features.

The contribution ratio (short for $r$ value) of the Conception feature is relatively larger in \FIG{fig_contributions} (a) since the target agent almost stands still and does not belong to any group with another agent in the scene.
This might lead the model to make predictions based on what he perceives through the Conception module more when both the ego and the group features are limited.
As for the family group example in \FIG{fig_group_method}, $r(\mathbf{f}^i_{\mathrm{group}})$ is more significant compared to other examples and $r(\mathbf{f}^i_{\mathrm{con}})$ also presents considerable contributions brought by the Conception feature, which aligns with the observation that there are relatively more neighboring agents around the Child(see \FIG{fig_group_method,fig_contributions} for comparisons).
Agent walking with a group member in the scene with no one else demonstrates a minor $r$ value of the Conception feature(\FIG{fig_contributions} (c)).
A similar phenomenon can be observed in the SDD dataset (\FIG{fig_contributions} (e) and (f)), and the $r$ value in the NBA dataset demonstrates an exceptionally high contribution of conception feature since the social interactions in NBA games are more abundant than those of other pedestrian scenes.

We also conduct ``intervention'' experiments \cite{wong2024socialcircle+} to further validate the cooperational property of the Group method and the Conception module as shown in \FIG{fig_manual_neighbor}.
In this work, we ``intervene'' in the prediction scenes from two different aspects.
Firstly, we add a manual ``unrelated'' neighbor to the target agent in the original zara1 scene (\FIG{fig_manual_neighbor} (a1)).
We can observe from the model predictions that the target agent tends to move downwards after placing this manual neighbor who moves right towards him (\FIG{fig_manual_neighbor} (a2)).
The attention value of the Conception module changes after proceeding with this ``intervention'', and the distinct rise of the value in the right partition is aligned with the position of the manual neighbor.
Accordingly, the $r(\mathbf{f}^i_{\mathrm{con}})$ becomes more prominent as more than two times the original one.
Furthermore, we add a manual ``related'' Group member to the target agent of an SDD littl0 \footnote{A subset of the SDD dataset.} scene (\FIG{fig_manual_neighbor} (b1)).
The attention value of the Conception module remains the same since the module could only perceive interactions with the ``unrelated'' agents.
Comparing \FIG{fig_manual_neighbor} (b1) and (b2), we could also observe a contribution ratio change between features that $r(\mathbf{f}^i_{\mathrm{group}})$ increases with $r(\mathbf{f}^i_{\mathrm{self}})$ dropping while $r(\mathbf{f}^i_{\mathrm{con}})$ stays almost the same as the original scene.

The visualization and ``intervention'' experiments above demonstrate how the Group method and the Conception module mutually contribute to the GPCC model's prediction performance and its effectiveness and explainability.

\section{Conclusion}

Inspired by how humans perceive other agents in the scene, this work proposes the GPCC model composed of the Group method, which could reveal grouping relations between pedestrians and the Conception module using visual information strategy and acoustic information strategy to perceive social interactions with agents out of the group.
Quantitative and qualitative analysis validate the effectiveness and the explainability of the proposed GPCC model and its components.

There are still limitations for further improvements.
The introduced Group method merely uses the observed trajectory coordinates to calculate the long-term distance kernel function between agents, and the group-split accuracy relies on the sequence length of historical trajectories.
However, humans can tell grouping relations within a short time.
Considering that little research is being done to study the grouping relations between agents and validate their effectiveness, such limitations might still need to be addressed to acquire better prediction performance.

{
    \small
    \bibliographystyle{ieeenat_fullname}
    \bibliography{ref.bib}

\begin{thebibliography}{46}
\providecommand{\natexlab}[1]{#1}
\providecommand{\url}[1]{\texttt{#1}}
\expandafter\ifx\csname urlstyle\endcsname\relax
  \providecommand{\doi}[1]{doi: #1}\else
  \providecommand{\doi}{doi: \begingroup \urlstyle{rm}\Url}\fi

\bibitem[Alahi et~al.(2016)Alahi, Goel, Ramanathan, Robicquet, Fei-Fei, and Savarese]{alahi2016social}
Alexandre Alahi, Kratarth Goel, Vignesh Ramanathan, Alexandre Robicquet, Li Fei-Fei, and Silvio Savarese.
\newblock Social lstm: Human trajectory prediction in crowded spaces.
\newblock In \emph{Proceedings of the IEEE conference on computer vision and pattern recognition}, pages 961--971, 2016.

\bibitem[Bae et~al.(2022)Bae, Park, and Jeon]{bae2022learning}
Inhwan Bae, Jin-Hwi Park, and Hae-Gon Jeon.
\newblock Learning pedestrian group representations for multi-modal trajectory prediction.
\newblock In \emph{European Conference on Computer Vision}, pages 270--289. Springer, 2022.

\bibitem[Bae et~al.(2024)Bae, Lee, and Jeon]{bae2024can}
Inhwan Bae, Junoh Lee, and Hae-Gon Jeon.
\newblock Can language beat numerical regression? language-based multimodal trajectory prediction.
\newblock \emph{arXiv preprint arXiv:2403.18447}, 2024.

\bibitem[Barata et~al.(2021)Barata, Nascimento, Lemos, and Marques]{barata2021sparse}
Catarina Barata, Jacinto~C. Nascimento, Jo{\~a}o~M. Lemos, and Jorge~S. Marques.
\newblock Sparse motion fields for trajectory prediction.
\newblock \emph{Pattern Recognition}, 110:\penalty0 107631, 2021.

\bibitem[Caesar et~al.(2020)Caesar, Bankiti, Lang, Vora, Liong, Xu, Krishnan, Pan, Baldan, and Beijbom]{caesar2020nuscenes}
Holger Caesar, Varun Bankiti, Alex~H Lang, Sourabh Vora, Venice~Erin Liong, Qiang Xu, Anush Krishnan, Yu Pan, Giancarlo Baldan, and Oscar Beijbom.
\newblock nuscenes: A multimodal dataset for autonomous driving.
\newblock In \emph{Proceedings of the IEEE/CVF conference on computer vision and pattern recognition}, pages 11621--11631, 2020.

\bibitem[Cao et~al.(2020)Cao, Wang, Duan, Zhang, Zhu, Huang, Tong, Xu, Bai, Tong, et~al.]{cao2020spectral}
Defu Cao, Yujing Wang, Juanyong Duan, Ce Zhang, Xia Zhu, Congrui Huang, Yunhai Tong, Bixiong Xu, Jing Bai, Jie Tong, et~al.
\newblock Spectral temporal graph neural network for multivariate time-series forecasting.
\newblock \emph{Advances in Neural Information Processing Systems}, 33:\penalty0 17766--17778, 2020.

\bibitem[Cao et~al.(2021)Cao, Li, Ma, and Tomizuka]{cao2021spectral}
Defu Cao, Jiachen Li, Hengbo Ma, and Masayoshi Tomizuka.
\newblock Spectral temporal graph neural network for trajectory prediction.
\newblock In \emph{2021 IEEE International Conference on Robotics and Automation (ICRA)}, pages 1839--1845. IEEE, 2021.

\bibitem[Chib and Singh(2024)]{chib2024lg}
Pranav~Singh Chib and Pravendra Singh.
\newblock Lg-traj: Llm guided pedestrian trajectory prediction.
\newblock \emph{arXiv preprint arXiv:2403.08032}, 2024.

\bibitem[Dong et~al.(2024)Dong, Wang, Zhou, Hua, and Sun]{dong2024recurrent}
Yonghao Dong, Le Wang, Sanping Zhou, Gang Hua, and Changyin Sun.
\newblock Recurrent aligned network for generalized pedestrian trajectory prediction.
\newblock \emph{arXiv preprint arXiv:2403.05810}, 2024.

\bibitem[Gupta et~al.(2018)Gupta, Johnson, Fei-Fei, Savarese, and Alahi]{gupta2018social}
Agrim Gupta, Justin Johnson, Li Fei-Fei, Silvio Savarese, and Alexandre Alahi.
\newblock Social gan: Socially acceptable trajectories with generative adversarial networks.
\newblock In \emph{Proceedings of the IEEE Conference on Computer Vision and Pattern Recognition}, pages 2255--2264, 2018.

\bibitem[Helbing and Molnar(1995)]{helbing1995social}
Dirk Helbing and Peter Molnar.
\newblock Social force model for pedestrian dynamics.
\newblock \emph{Physical review E}, 51\penalty0 (5):\penalty0 4282, 1995.

\bibitem[Kim et~al.(2024)Kim, Chi, Lim, Ramani, Kim, and Kim]{kim2024higher}
Sungjune Kim, Hyung-gun Chi, Hyerin Lim, Karthik Ramani, Jinkyu Kim, and Sangpil Kim.
\newblock Higher-order relational reasoning for pedestrian trajectory prediction.
\newblock In \emph{Proceedings of the IEEE/CVF Conference on Computer Vision and Pattern Recognition}, pages 15251--15260, 2024.

\bibitem[Kothari et~al.(2021)Kothari, Kreiss, and Alahi]{kothari2020human}
Parth Kothari, Sven Kreiss, and Alexandre Alahi.
\newblock Human trajectory forecasting in crowds: A deep learning perspective.
\newblock \emph{IEEE Transactions on Intelligent Transportation Systems}, 23\penalty0 (7):\penalty0 7386--7400, 2021.

\bibitem[Lee et~al.(2022)Lee, Sohn, Moon, Yoon, Kapadia, and Pavlovic]{lee2022muse}
Mihee Lee, Samuel~S Sohn, Seonghyeon Moon, Sejong Yoon, Mubbasir Kapadia, and Vladimir Pavlovic.
\newblock Muse-vae: Multi-scale vae for environment-aware long term trajectory prediction.
\newblock In \emph{Proceedings of the IEEE/CVF Conference on Computer Vision and Pattern Recognition}, pages 2221--2230, 2022.

\bibitem[Lerner et~al.(2007)Lerner, Chrysanthou, and Lischinski]{lerner2007crowds}
Alon Lerner, Yiorgos Chrysanthou, and Dani Lischinski.
\newblock Crowds by example.
\newblock \emph{Computer Graphics Forum}, 26\penalty0 (3):\penalty0 655--664, 2007.

\bibitem[Li et~al.(2022)Li, Yang, and Sun]{li2022intention}
Cunyan Li, Hua Yang, and Jun Sun.
\newblock Intention-interaction graph based hierarchical reasoning networks for human trajectory prediction.
\newblock \emph{IEEE Transactions on Multimedia}, 2022.

\bibitem[Li et~al.(2021)Li, Zhou, Yi, and Gall]{li2021spatial}
Shijie Li, Yanying Zhou, Jinhui Yi, and Juergen Gall.
\newblock Spatial-temporal consistency network for low-latency trajectory forecasting.
\newblock In \emph{Proceedings of the IEEE/CVF International Conference on Computer Vision (ICCV)}, pages 1940--1949, 2021.

\bibitem[Liang et~al.(2020)Liang, Jiang, and Hauptmann]{liang2020simaug}
Junwei Liang, Lu Jiang, and Alexander Hauptmann.
\newblock Simaug: Learning robust representations from simulation for trajectory prediction.
\newblock In \emph{Proceedings of the European conference on computer vision (ECCV)}, 2020.

\bibitem[Linou et~al.(2016)Linou, Linou, and de~Boer]{linou2016nba}
Kostya Linou, Dzmitryi Linou, and Martijn de Boer.
\newblock Nba player movements.
\newblock https://github.com/linouk23/NBA-Player-Movements, 2016.

\bibitem[Liu et~al.(2024)Liu, Ye, Wang, Li, Sheng, and Yao]{liu2024uncertainty}
Yao Liu, Zesheng Ye, Rui Wang, Binghao Li, Quan~Z Sheng, and Lina Yao.
\newblock Uncertainty-aware pedestrian trajectory prediction via distributional diffusion.
\newblock \emph{Knowledge-Based Systems}, page 111862, 2024.

\bibitem[Luber et~al.(2010)Luber, Stork, Tipaldi, and Arras]{luber2010people}
Matthias Luber, Johannes~A Stork, Gian~Diego Tipaldi, and Kai~O Arras.
\newblock People tracking with human motion predictions from social forces.
\newblock In \emph{2010 IEEE international conference on robotics and automation}, pages 464--469. IEEE, 2010.

\bibitem[Maeda and Ukita(2023)]{maeda2023fast}
Takahiro Maeda and Norimichi Ukita.
\newblock Fast inference and update of probabilistic density estimation on trajectory prediction.
\newblock In \emph{Proceedings of the IEEE/CVF International Conference on Computer Vision}, pages 9795--9805, 2023.

\bibitem[Mangalam et~al.(2021)Mangalam, An, Girase, and Malik]{mangalam2020s}
Karttikeya Mangalam, Yang An, Harshayu Girase, and Jitendra Malik.
\newblock From goals, waypoints \& paths to long term human trajectory forecasting.
\newblock In \emph{Proceedings of the IEEE/CVF International Conference on Computer Vision}, pages 15233--15242, 2021.

\bibitem[Mehran et~al.(2009)Mehran, Oyama, and Shah]{mehran2009abnormal}
Ramin Mehran, Alexis Oyama, and Mubarak Shah.
\newblock Abnormal crowd behavior detection using social force model.
\newblock In \emph{2009 IEEE Conference on Computer Vision and Pattern Recognition}, pages 935--942. IEEE, 2009.

\bibitem[Mohamed et~al.(2020)Mohamed, Qian, Elhoseiny, and Claudel]{mohamed2020social}
Abduallah Mohamed, Kun Qian, Mohamed Elhoseiny, and Christian Claudel.
\newblock Social-stgcnn: A social spatio-temporal graph convolutional neural network for human trajectory prediction.
\newblock In \emph{Proceedings of the IEEE/CVF Conference on Computer Vision and Pattern Recognition}, pages 14424--14432, 2020.

\bibitem[Pei et~al.(2019)Pei, Qi, Zhang, Ma, and Yang]{pei2019human}
Zhao Pei, Xiaoning Qi, Yanning Zhang, Miao Ma, and Yee-Hong Yang.
\newblock Human trajectory prediction in crowded scene using social-affinity long short-term memory.
\newblock \emph{Pattern Recognition}, 93:\penalty0 273--282, 2019.

\bibitem[Pellegrini et~al.(2009)Pellegrini, Ess, Schindler, and Van~Gool]{pellegrini2009youll}
Stefano Pellegrini, Andreas Ess, Konrad Schindler, and Luc Van~Gool.
\newblock You'll never walk alone: Modeling social behavior for multi-target tracking.
\newblock In \emph{2009 IEEE 12th International Conference on Computer Vision}, pages 261--268. IEEE, 2009.

\bibitem[Robicquet et~al.(2016)Robicquet, Sadeghian, Alahi, and Savarese]{robicquet2016learning}
Alexandre Robicquet, Amir Sadeghian, Alexandre Alahi, and Silvio Savarese.
\newblock Learning social etiquette: Human trajectory understanding in crowded scenes.
\newblock In \emph{European conference on computer vision}, pages 549--565. Springer, 2016.

\bibitem[Saadatnejad et~al.(2022)Saadatnejad, Ju, and Alahi]{saadatnejad2022pedestrian}
Saeed Saadatnejad, Yi~Zhou Ju, and Alexandre Alahi.
\newblock Pedestrian 3d bounding box prediction.
\newblock \emph{arXiv preprint arXiv:2206.14195}, 2022.

\bibitem[Salzmann et~al.(2020)Salzmann, Ivanovic, Chakravarty, and Pavone]{salzmann2020trajectron}
Tim Salzmann, Boris Ivanovic, Punarjay Chakravarty, and Marco Pavone.
\newblock Trajectron++: Dynamically-feasible trajectory forecasting with heterogeneous data.
\newblock In \emph{Proceedings of the European conference on computer vision (ECCV)}, pages 683--700. Springer, 2020.

\bibitem[Shi et~al.(2023)Shi, Wang, Long, Zhou, Tang, Zheng, and Hua]{shi2023representing}
Liushuai Shi, Le Wang, Chengjiang Long, Sanping Zhou, Wei Tang, Nanning Zheng, and Gang Hua.
\newblock Representing multimodal behaviors with mean location for pedestrian trajectory prediction.
\newblock \emph{IEEE Transactions on Pattern Analysis and Machine Intelligence}, 2023.

\bibitem[Su et~al.(2022)Su, Du, Li, Li, Liang, Hua, and Zhou]{su2022trajectory}
Yuchao Su, Jie Du, Yuanman Li, Xia Li, Rongqin Liang, Zhongyun Hua, and Jiantao Zhou.
\newblock Trajectory forecasting based on prior-aware directed graph convolutional neural network.
\newblock \emph{IEEE Transactions on Intelligent Transportation Systems}, pages 1--13, 2022.

\bibitem[Vaswani et~al.(2017)Vaswani, Shazeer, Parmar, Uszkoreit, Jones, Gomez, Kaiser, and Polosukhin]{vaswani2017attention}
Ashish Vaswani, Noam Shazeer, Niki Parmar, Jakob Uszkoreit, Llion Jones, Aidan~N Gomez, {\L}ukasz Kaiser, and Illia Polosukhin.
\newblock Attention is all you need.
\newblock In \emph{Advances in neural information processing systems}, pages 5998--6008, 2017.

\bibitem[Vemula et~al.(2017)Vemula, Muelling, and Oh]{vemula2017modeling}
Anirudh Vemula, Katharina Muelling, and Jean Oh.
\newblock Modeling cooperative navigation in dense human crowds.
\newblock In \emph{2017 IEEE International Conference on Robotics and Automation (ICRA)}, pages 1685--1692. IEEE, 2017.

\bibitem[Von~Helmholtz(1867)]{von1867handbuch}
Hermann Von~Helmholtz.
\newblock \emph{Handbuch der physiologischen Optik}.
\newblock Voss, 1867.

\bibitem[Wang et~al.(2023)Wang, Liu, Wang, Wang, Wang, and Mcloone]{wang2022seem}
Dafeng Wang, Hongbo Liu, Naiyao Wang, Yiyang Wang, Hua Wang, and Sean Mcloone.
\newblock Seem: a sequence entropy energy-based model for pedestrian trajectory all-then-one prediction.
\newblock \emph{IEEE transactions on pattern analysis and machine intelligence}, 45\penalty0 (1):\penalty0 1070--1086, 2023.

\bibitem[Wong et~al.(2023{\natexlab{a}})Wong, Xia, Peng, and You]{wong2023another}
Conghao Wong, Beihao Xia, Qinmu Peng, and Xinge You.
\newblock Another vertical view: A hierarchical network for heterogeneous trajectory prediction via spectrums.
\newblock \emph{arXiv preprint arXiv:2304.05106}, 2023{\natexlab{a}}.

\bibitem[Wong et~al.(2023{\natexlab{b}})Wong, Xia, Peng, Yuan, and You]{wong2021msn}
Conghao Wong, Beihao Xia, Qinmu Peng, Wei Yuan, and Xinge You.
\newblock Msn: multi-style network for trajectory prediction.
\newblock \emph{IEEE Transactions on Intelligent Transportation Systems}, 24:\penalty0 9751 -- 9766, 2023{\natexlab{b}}.

\bibitem[Wong et~al.(2024{\natexlab{a}})Wong, Xia, Zou, Wang, and You]{wong2023socialcircle}
Conghao Wong, Beihao Xia, Ziqian Zou, Yulong Wang, and Xinge You.
\newblock Socialcircle: Learning the angle-based social interaction representation for pedestrian trajectory prediction.
\newblock In \emph{Proceedings of the IEEE/CVF Conference on Computer Vision and Pattern Recognition}, pages 19005--19015, 2024{\natexlab{a}}.

\bibitem[Wong et~al.(2024{\natexlab{b}})Wong, Xia, Zou, and You]{wong2024socialcircle+}
Conghao Wong, Beihao Xia, Ziqian Zou, and Xinge You.
\newblock Socialcircle+: Learning the angle-based conditioned interaction representation for pedestrian trajectory prediction.
\newblock \emph{arXiv preprint arXiv:2409.14984}, 2024{\natexlab{b}}.

\bibitem[Xu et~al.(2022{\natexlab{a}})Xu, Li, Ni, Zhang, and Chen]{xu2022groupnet}
Chenxin Xu, Maosen Li, Zhenyang Ni, Ya Zhang, and Siheng Chen.
\newblock Groupnet: Multiscale hypergraph neural networks for trajectory prediction with relational reasoning.
\newblock In \emph{Proceedings of the IEEE/CVF Conference on Computer Vision and Pattern Recognition (CVPR)}, pages 6498--6507, 2022{\natexlab{a}}.

\bibitem[Xu et~al.(2022{\natexlab{b}})Xu, Mao, Zhang, and Chen]{xu2022remember}
Chenxin Xu, Weibo Mao, Wenjun Zhang, and Siheng Chen.
\newblock Remember intentions: Retrospective-memory-based trajectory prediction.
\newblock In \emph{Proceedings of the IEEE/CVF Conference on Computer Vision and Pattern Recognition (CVPR)}, pages 6488--6497, 2022{\natexlab{b}}.

\bibitem[Xu et~al.(2023)Xu, Tan, Tan, Chen, Wang, Wang, and Wang]{xu2023eqmotion}
Chenxin Xu, Robby~T Tan, Yuhong Tan, Siheng Chen, Yu~Guang Wang, Xinchao Wang, and Yanfeng Wang.
\newblock Eqmotion: Equivariant multi-agent motion prediction with invariant interaction reasoning.
\newblock In \emph{Proceedings of the IEEE/CVF Conference on Computer Vision and Pattern Recognition}, pages 1410--1420, 2023.

\bibitem[Yuan et~al.(2021)Yuan, Weng, Ou, and Kitani]{yuan2021agentformer}
Ye Yuan, Xinshuo Weng, Yanglan Ou, and Kris~M. Kitani.
\newblock Agentformer: Agent-aware transformers for socio-temporal multi-agent forecasting.
\newblock In \emph{Proceedings of the IEEE/CVF International Conference on Computer Vision (ICCV)}, pages 9813--9823, 2021.

\bibitem[Yue et~al.(2022)Yue, Manocha, and Wang]{yue2022human}
Jiangbei Yue, Dinesh Manocha, and He Wang.
\newblock Human trajectory prediction via neural social physics.
\newblock In \emph{European Conference on Computer Vision}, pages 376--394. Springer, 2022.

\bibitem[Zhang et~al.(2022)Zhang, Xue, Zhang, Zheng, and Ouyang]{zhang2020social}
Pu Zhang, Jianru Xue, Pengfei Zhang, Nanning Zheng, and Wanli Ouyang.
\newblock Social-aware pedestrian trajectory prediction via states refinement lstm.
\newblock \emph{IEEE transactions on pattern analysis and machine intelligence}, 44\penalty0 (5):\penalty0 2742--2759, 2022.

\end{thebibliography}
}


\maketitlesupplementary

\appendix
\setcounter{table}{0}
\setcounter{figure}{0}
\renewcommand{\thetable}{S\arabic{table}}
\renewcommand{\thefigure}{S\arabic{figure}}

\begin{figure*}[t]
    \centering
    \includegraphics[width=1.0\linewidth]{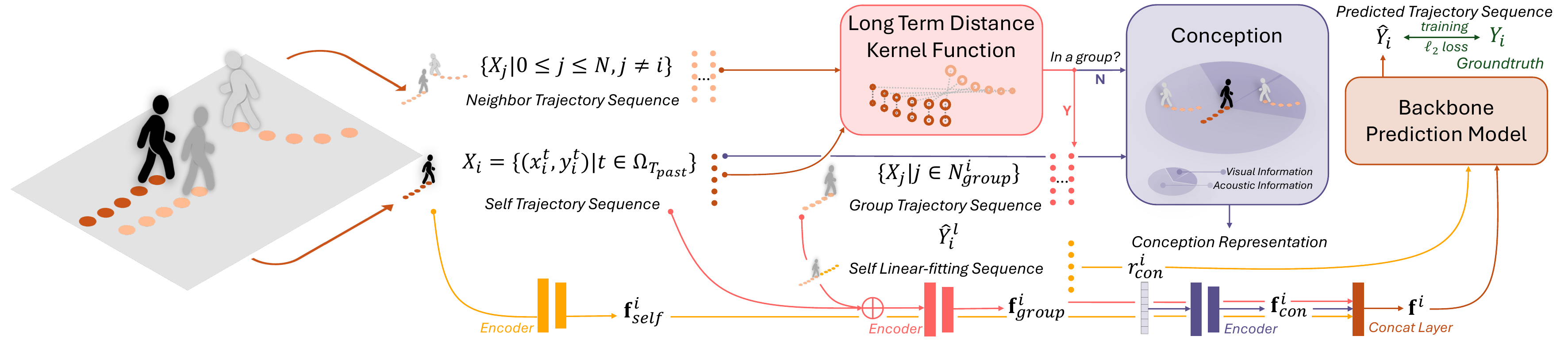}
    \caption{
        More detailed version of the schema of the modeling process of the proposed GPCC model and the computation pipeline of the Group method and the Conception module.
    }
    \label{fig_method_long}
\end{figure*}

\section{Additional Quantitative Analysis}

We only report the GPCC model's performance with part of the quantitative results due to page limitations.
As the qualitative analysis shown in the main manuscript, the proposed GPCC model presents the capability to handle different prediction scenes.
This section further validates the model's effectiveness by presenting additional quantitative analysis of the datasets qualitatively analyzed in the manuscript.

\subsection{Experimental Configurations}

\textbf{NBA}~comprises trajectories of both players and basketball captured by SportVU tracking systems during NBA games.
Following the methodology proposed by Xu et al.\cite{xu2022groupnet,xu2022remember}, we set the parameters to be $\left\{n_p, n_f, T\right\} = \left\{5, 10, 0.4\mathrm{s}\right\}$, randomly selecting approximately 50000 samples (ego trajectories), with 65\% allocated for training, 25\% for testing, and 10\% for validation.

\textbf{nuScenes}~contains 1000 driving scenes collected in the urban area of Boston and Singapore.
Each scene is 20 seconds long and is annotated at a rate of 2 frames per second.
In the manuscript, we only use the two-dimensional trajectories of vehicles to evaluate our GPCC model.
We follow the methodology proposed by \cite{lee2022muse} of $\left\{n_p = 4, n_f = 12, T = 0.5 \mathrm{s}\right\}$, and training strategy proposed by \cite{saadatnejad2022pedestrian} of using 550 scenes to train, 150 scenes to validate, and the other 150 scenes to test.

\begin{table}[htbp]
    \centering
    \footnotesize
    \begin{tabular}{|lcc|}
        \toprule
        Model & $n_p = 5$ & $n_f = 10$ \\

        \midrule
        \SOCIALLSTM\SOCIALLSTMCITE~(2016) & 0.88/1.53 & 1.79/3.16 \\
        \SOCIALGAN\SOCIALGANCITE~(2018) & 0.85/1.36 & 1.62/2.51 \\
        \STGCNN\STGCNNCITE~(2020) & 0.75/0.99 & 1.59/2.37 \\
        GroupNet+NMMP\cite{xu2022groupnet}~(2022) & 0.69/1.08 & 1.25/1.80 \\
        GroupNet+CVAE\cite{xu2022groupnet}~(2022) & \textbf{0.62}/0.95 & \textbf{1.13}/1.69 \\
        \SCMODEL\SCCITE~(2024) & \underline{0.67}/\underline{0.90} & \underline{1.18}/\textbf{1.46} \\
        
        \midrule
        GPCC~\textbf{(Ours)} & \textbf{0.62}/\textbf{0.87} & 1.19/\underline{1.58} \\

        \bottomrule
    \end{tabular}
    \caption{
        Comparisons on NBA under $\{n_p, n_f, T\} = \{5, 10, 0.4\mathrm{s}\}$.
 Metrics are ``ADE/FDE'' in meters under \emph{best-of-20} on ${5, 10}$ future steps, and lower ADE and FDE indicate better performance.
    }
    \label{tab_nba_comparisons}
\end{table}

\begin{table}[htbp]
    \centering
    \footnotesize
    \begin{tabular}{|lcc|}
        \toprule
        Model & \emph{best-of-5} & \emph{best-of-10} \\

        \midrule
        Trajectron++\TRAJECTRONPPCITE~(2020) & 3.14/7.45 & 2.46/5.65 \\
        \YNET\YNETCITE~(2020) & 2.46/5.15 & 1.88/3.47 \\
        \AGENTFORMER\AGENTFORMERCITE~(2021) & 1.59/3.14 & 1.30/2.47 \\
        \EVMODEL\EVCITE~(2023) & 1.46/3.18 & 1.15/2.37 \\
        \SCMODEL\SCCITE~(2024) & 1.44/3.10 & 1.13/2.30 \\
        \MUSEVAE\MUSEVAECITE~(2022) & \underline{1.38}/\textbf{2.90} & \underline{1.09}/\textbf{2.10} \\
        
        \midrule
        GPCC~\textbf{(Ours)} & \textbf{1.33}/\underline{2.94} & \textbf{1.08}/\underline{2.27} \\

        \bottomrule
    \end{tabular}
    \caption{
        Comparisons on nuScenes under $\{n_p, n_f, T\} = \{4, 12, 0.5\mathrm{s}\}$.
        Metrics are ``ADE/FDE'' in meters under \emph{best-of-$k$} ($k={5, 10}$) and lower values indicate better prediction performance.
    }
    \label{tab_nuscenes_comparisons}
\end{table}

\subsection{Analysis}

NBA players on the NBA dataset interact with each other differently compared with ETH-UCY and SDD.
In \TABLE{tab_nba_comparisons}, we can observe that the GPCC model outperforms other state-of-the-art methods when $n_p = 5$ and reaches the second best in FDE when $n_p=10$.
The results of NBA validate the GPCC model's capability of modeling different social interactions.

We only consider trajectories of vehicles only on nuScenes.
Interactions between vehicles could differ entirely from those between pedestrians, and there might not be such \emph{group} relations between them.
This property of nuScenes are discussed in Sec. 4.5.
However, the GPCC model still gained a considerable prediction performance, as shown in \TABLE{tab_nuscenes_comparisons}.

\begin{table*}[htbp]
    \centering
    \footnotesize
    \begin{tabular}{|c|cc|cc|cc|cc|cc|ccccccccc}
        \toprule
        \multirow{2}{*}{ID} & \multirow{2}{*}{Group} & \multirow{2}{*}{Conception} &
        \multicolumn{2}{|c|}{NBA} & \multicolumn{2}{|c|}{$\Delta$NBA} & \multicolumn{2}{|c|}{nuScenes} & \multicolumn{2}{|c|}{$\Delta$nuScenes} \\
        &&& $n_p = 5$ & $n_f = 10$ & $n_p = 5$ & $n_f = 10$ & \emph{best-of-5} & \emph{best-of-10} & \emph{best-of-5} & \emph{best-of-10} \\

        \midrule
        v1 & \multicolumn{2}{|l|}{~~~~~$\bullet$~~~~~~~~~~~~~~~~~$\circ$} 
         & 0.62/0.87 & 1.19/1.58 & 0.0\%/0.0\% & 0.0\%/0.0\% &1.38/3.05&1.10/2.30&3.8\%/3.7\% & 1.9\%/0.9\%\\
        v2 & \multicolumn{2}{|l|}{~~~~~$\circ$~~~~~~~~~~~~~~~~~$\bullet$} 
        & 0.66/0.94 & 1.25/1.68 & 6.5\%/8.0\% & 5.0\%/6.3\% &1.34/2.95&1.07/2.27&0.8\%/0.3\% & -0.9\%/0.0\%\\
        v3 & \multicolumn{2}{|l|}{~~~~~$\circ$~~~~~~~~~~~~~~~~~$\circ$} 
       & 0.70/1.03 & 1.34/1.86 & 12.9\%/18.4\% & 12.6\%/17.7\% &1.34/2.95&1.09/2.28&0.8\%/0.3\% & 0.9\%/0.4\%\\

        \midrule
        v0 & \multicolumn{2}{|l|}{~~~~~$\bullet$~~~~~~~~~~~~~~~~~$\bullet$} 
         & 0.62/0.87 & 1.19/1.58 & 0.0\%/0.0\% & 0.0\%/0.0\% &1.33/2.94 & 1.08/2.27&0.0\%/0.0\% & 0.0\%/0.0\%\\

        \bottomrule
    \end{tabular}
    \caption{
      Ablation studies on NBA and nuScenes.
      ``$\bullet$'' means using the corresponding method or module and ``$\circ$'' indicates the opposite.
      ``ID'' represents a different variation index of the proposed GPCC model, and ``$\Delta{\mathrm{NBA, nuScenes}}$'' indicates the percentage of \textbf{performance drops} compared to the full GPCC model(v0).
    }
    \label{tab_nbanuscenes_variations}
  \end{table*}

\begin{table}[htbp]
    \centering
    \footnotesize
    \begin{tabular}{|c|cc|cc|ccccccccccccccc}
        \toprule
        ID & Group & Conception &
        SDD &
        $\Delta$SDD \\

        \midrule
        v1 & \multicolumn{2}{|l|}{~~~~~$\bullet$~~~~~~~~~~~~~~~~~$\circ$} 
         & 6.44/10.34 & 0.8\%/1.7\%\\
        v2 & \multicolumn{2}{|l|}{~~~~~$\circ$~~~~~~~~~~~~~~~~~$\bullet$} 
        & 6.46/10.38 & 1.1\%/2.1\%\\
        v3 & \multicolumn{2}{|l|}{~~~~~$\circ$~~~~~~~~~~~~~~~~~$\circ$} 
       & 6.40/10.17 & 0.2\%/0\%\\

        \midrule
        v0 & \multicolumn{2}{|l|}{~~~~~$\bullet$~~~~~~~~~~~~~~~~~$\bullet$} 
         & 6.39/10.17 & 0.0\%/0.0\%\\

        \bottomrule
    \end{tabular}
    \caption{
      Ablation studies on SDD.
      ``$\bullet$'' means using the corresponding method or module and ``$\circ$'' indicates the opposite.
      ``ID'' represents a different variation index of the proposed GPCC model, and ``$\Delta$SDD'' indicates the percentage of \textbf{performance drops} compared to the full GPCC model(v0).
    }
    \label{tab_sdd_variations}
  \end{table}

Although results in \TABLE{tab_sdd_variations} are relatively minor compared to those in Tab. 1, we could still see a performance drop of larger ADE and FDE at v1,v2 and v3.
We could observe that agents in SDD tend to move and behave independently, and there seems to be less interaction between agents from the relatively high eye-bird view compared to scenes in ETH-UCY, and this might be the reason for the smaller contributions of the Group method and Conception module of the proposed GPCC model when evaluating on the SDD dataset.

In \TABLE{tab_nbanuscenes_variations}, we could observe that the performance drops the most when disabling both the Group method and Conception module on the NBA dataset.
However, the prediction accuracy when disabling the Conception module only (v1) reaches the same level as the original GPCC model (discussed in Sec. 4.5).
The results demonstrate an opposite performance change on the nuScenes dataset.
The prediction performance drops when disabling the Conception module (v1) or both of them (v3) and keeps the same level as the original GPCC model 
when disabling the Group method only (v2).
The reason for this phenomenon might be common sense that vehicles moving on the road are not in distinct groups. 
It might lead the model to learn information through group relations between cars if we use the Group method on nuScenes.

\section{Additional Analysis of Time Efficiency}

Pedestrian trajectory prediction task requires low-latency prediction performance to be integrated into corresponding applications, \EG, autonomous driving.
We also use Apple Mac Studio (M2 Max) to evaluate the time efficiency of different methods, as shown in Sec. 4.4.
Considering the practical amount of pedestrians moving in a multi-agent scene, we evaluate the average inference time of 100 target agents (batchsize 100) using different methods.

The proposed GPCC model demonstrates considerable inference speed compared with other methods using Transformer as part of backbone prediction model\cite{wong2023socialcircle} (32ms with batchsize 100),\cite{wong2024socialcircle+} (96ms with batchsize 100).
With the inference time already satisfying handling 99 more agents between two adjacent intervals, the model should meet the low-latency requirement of the trajectory prediction task\cite{li2021spatial}.

\section{Additional Analysis of FOV Partitions}

As represented in \TABLE{tab_ethucy_fov,tab_nba_fov}, we first conduct variation experiments on pedestrian dataset ETH-UCY and game dataset NBA to observe how the performance of the proposed GPCC model vary with FOV angle $\theta_\mathrm{FOV}$.

\begin{table*}[htbp]
    \centering
    \footnotesize
    \begin{tabular}{|c|c|ccccccc|ccccccccccc}
        \toprule
        ID & $\theta_\mathrm{FOV}$ &
        eth &
        hotel &
        univ &
        zara1 &
        zara2 &
        ETH-UCY &
        $\Delta$ETH-UCY \\

        \midrule
        v4 & $0^\circ$ & 0.26/0.40 & 0.10/0.15 & 0.26/0.47 & 0.18/0.30 & 0.13/0.25 & 0.19/0.31 & 5.6\%/6.9\%\\
        v5 & $90^\circ$ & 0.25/0.38 & 0.11/0.16 & 0.26/0.45 & 0.18/0.30 & 0.14/0.23 & 0.19/0.30 & 5.6\%/3.4\%\\
        v6 & $135^\circ$ & 0.26/0.39 & 0.10/0.16 & 0.26/0.46 & 0.18/0.30 & 0.14/0.23 & 0.19/0.31 & 5.6\%/6.9\%\\
        v7 & $270^\circ$ & 0.26/0.41 & 0.11/0.17 & 0.26/0.46 & 0.18/0.31 & 0.14/0.23 & 0.19/0.32 & 5.6\%/10.3\%\\
        v8 & $360^\circ$ & 0.26/0.40 & 0.10/0.15 & 0.27/0.49 & 0.18/0.31 & 0.14/0.23 & 0.19/0.32 & 5.6\%/10.3\%\\

        \midrule
        v0 & $180^\circ$ & 0.25/0.38 & 0.10/0.15 & 0.25/0.44 & 0.17/0.28 & 0.13/0.22 & 0.18/0.29 & 0.0\%/0.0\%\\

        \bottomrule
    \end{tabular}
    \caption{
      FOV angle $\theta_{\mathrm{FOV}}$ analysis on EHT-UCY dataset.
      Variation ``ID'' is continuous from Tab. 3.
    }
    \label{tab_ethucy_fov}
  \end{table*}

\begin{table}[htbp]
    \centering
    \footnotesize
    \begin{tabular}{|c|c|cc|cc|cccccccccccccc}
        \toprule
        \multirow{2}{*}{ID} & \multirow{2}{*}{$\theta_\mathrm{FOV}$} & \multicolumn{2}{|c|}{NBA} & \multicolumn{2}{|c|}{$\Delta$NBA} \\
        && $n_p = 5$ & $n_f = 10$ & $n_p = 5$ & $n_f = 10$ \\

        \midrule
        v4 & $0^\circ$ & 0.62/0.86 & 1.19/1.57 &  0.0\%/-1.1\% & -0.8\%/-1.3\% \\
        v5 & $90^\circ$ & 0.64/0.89 & 1.22/1.62 &  3.2\%/2.3\% & 2.5\%/1.9\% \\
        v6 & $135^\circ$ & 0.63/0.88 & 1.20/1.60 &  1.6\%/1.1\% & 0.8\%/1.3\% \\
        v7 & $270^\circ$ & 0.62/0.86 & 1.19/1.58 &  0.0\%/-1.1\% & 0.0\%/0.0\% \\
        v8 & $360^\circ$ & 0.62/0.87 & 1.20/1.59 &  0.0\%/0.0\% & 0.8\%/0.6\% \\

        \midrule
        v0 & $180^\circ$ & 0.62/0.87 & 1.19/1.58 & 0.0\%/0.0\% & 0.0\%/0.0\%\\

        \bottomrule
    \end{tabular}
    \caption{
      FOV angle $\theta_{\mathrm{FOV}}$ analysis on NBA dataset.
      Variation ``ID'' is continuous from \FIG{tab_nbanuscenes_variations}.
    }
    \label{tab_nba_fov}
  \end{table}

\begin{figure*}[tbph]
    \centering
    \includegraphics[width=1.0\linewidth]{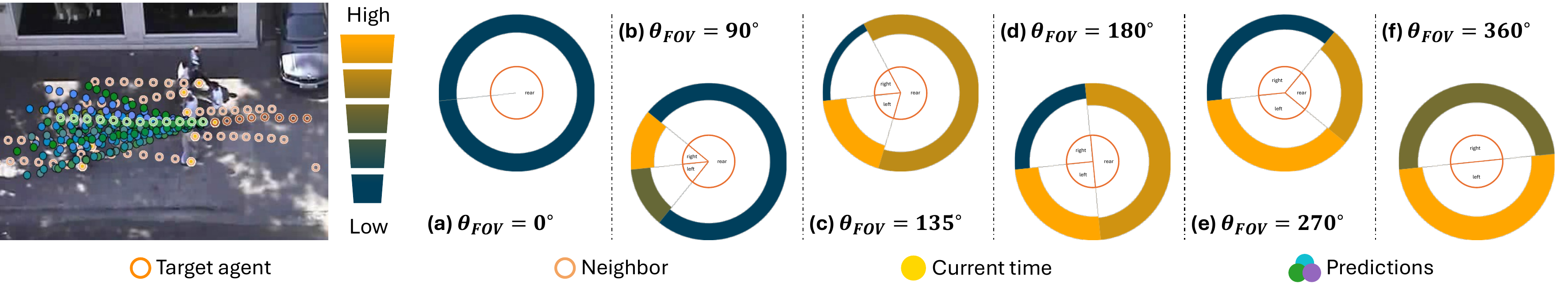}
    \caption{
        Visualization of attention value in the form of concentric fan chart varying with FOV angle $\theta_\mathrm{FOV}$.
        The model pays more attention to wider regions, whose color tends to be yellow accordingly.
    }
    \label{fig_fov_change}
\end{figure*}

In \FIG{tab_ethucy_fov}, the best performance comes at the original model(v0) whose $\theta_\mathrm{FOV}$ is set to be $180^\circ$.
This result is in line with the previous study \cite{von1867handbuch} that a human's single-eye FOV angle is around $150^\circ$ and the combined FOV from both eyes reaches about $200^\circ$, which is why we chose to set the $\theta_\mathrm{FOV} = 180^\circ$ at the first place.
Although we can observe that the performance of the GPCC model drops little from the overall results on the whole ETH-UCY dataset, the prediction performance on specific subsets such as univ, zara1, and zara2 drops relatively more than the other subsets.
This might be aroused that there are more social interactions in these subsets and the change of $\theta_\mathrm{FOV}$ modifies how the Conception module perceives social interactions with other agents.

Things are getting more interesting in the NBA dataset shown in \TABLE{tab_nba_fov}.
It can be observed that the prediction performance is becoming better from $\theta_\mathrm{FOV} = 90^\circ$ to $\theta_\mathrm{FOV} = 360^\circ$ (v5, v6, v7 and v8).
NBA players might need to spread their attention to a broader FOV to gain more information on the court and make decisions of movements based on this information.
Furthermore, when the $\theta_\mathrm{FOV} = 0^\circ$, the Conception module perceives interactions all the same by only considering the distance factor of other agents, which might lead to surprisingly minor improvements in the prediction performance.

We further visualize the attention value of the Conception module at different FOV angle settings with different concentric fan charts as shown in \FIG{fig_fov_change}.
By comparing charts when $\theta_\mathrm{FOV} = 90^\circ$ and $\theta_\mathrm{FOV} = 135^\circ$ (\FIG{fig_fov_change} (b) and (c)), we can observe a change of relative attention value in right and left partitions.
When using a wider FOV angle, agents divided into rear partitions at a narrower FOV angle can be included into left or right partitions so that they can be paid more attention.

\section{Additional Analysis of Choosing the Long-term Distance Threshold}

\begin{figure*}[t]
    \centering
    \includegraphics[width=1.0\linewidth]{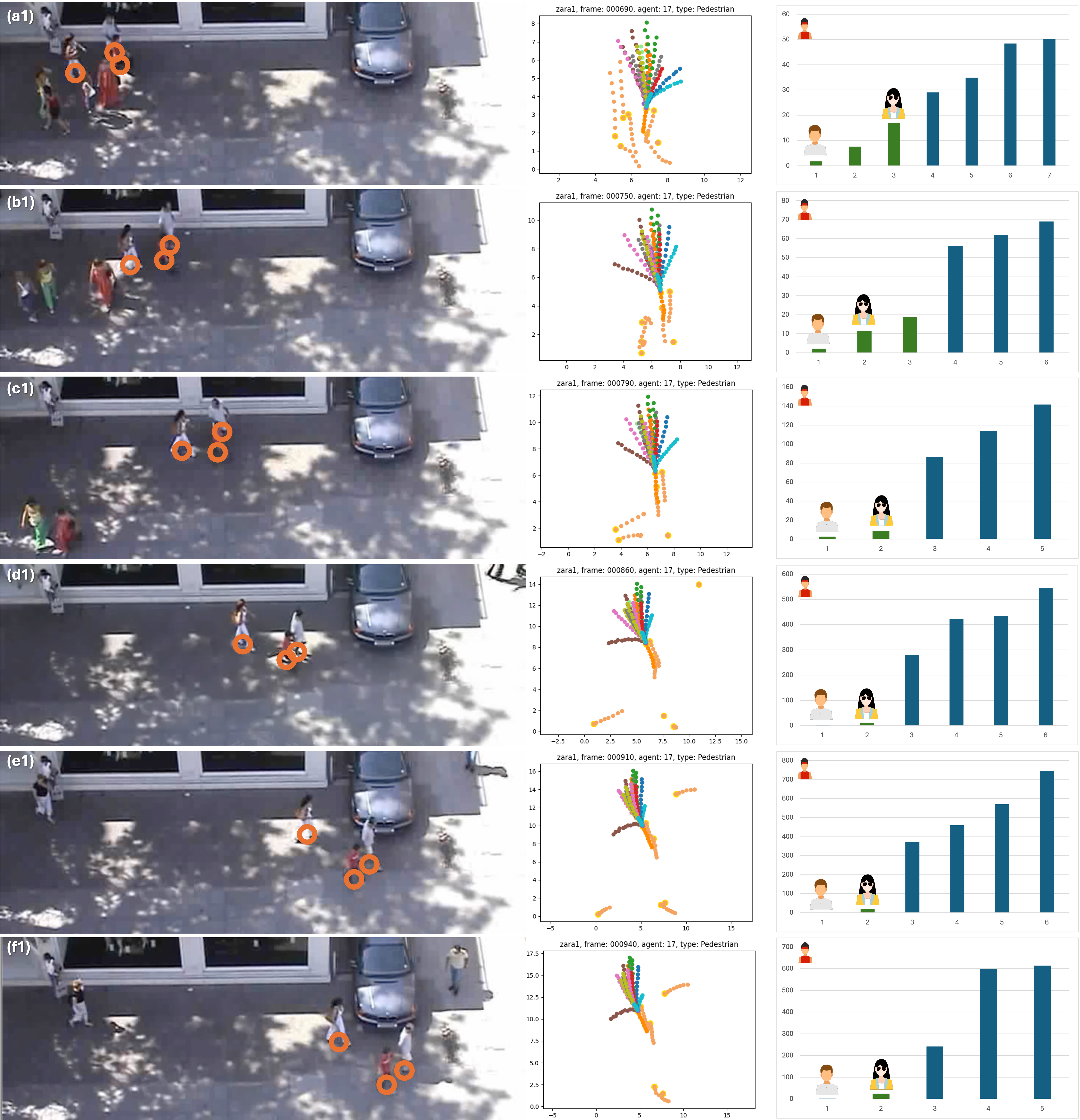}
    \caption{
        Illustration of how we design the threshold $d_\mathrm{m}$ of the long-term distance kernel function based on the sum of distance in bar charts corresponding to Fig. 3.
    }
    \label{fig_threshold}
\end{figure*}

In the main manuscript, we introduce the Group method and its core component, long-term distance kernel function $K(\cdot)$.
When calculating the long-term distance kernel function, we mention the threshold to determine whether the agent belongs to the same group as the target agent without detailedly introducing how to choose the threshold $d_\mathrm{m}$ due to page limitations.
This section demonstrates how we determine the kernel function's threshold $d_\mathrm{m}$.

As shown in the manuscript, we also use the family group here as an example,  considering Child as the target agent.
We also use this example because this family example includes diverse scenes from multiple neighboring agents to no one else around.
Members of the group (Mother, Father, and Child) behave differently according to their own will.
Overall, Child seems to be talking with Father all the way from the Zara store to the other side of the road.
Mother walks behind Child at a relatively larger distance compared to Father.
In \FIG{fig_threshold} (c1), we can observe that Father turns around to Mother to say something (at frame No.790), which further validates their grouping relations. 

The long-term distance value of each agent is shown in the bar charts on the right.
In \FIG{fig_threshold} (c1),(d1),(e1), and (f1), the long-term distance of Father-Child or Mother-Child (marked in deep green) is distinctively lower than other agents (marked in deep blue).
However, in \FIG{fig_threshold} (a1), we can observe that the long-term distance of Mother ranks third among all neighboring agents of the target agent Child while Father is still at the top of the list.
When plenty of agents exist in the scene, the long-term distance can be near each other when the time window used to calculate the distance sum is relatively short.
Although an ``unrelated'' neighbor is classified as a group member, the trajectories seem reasonable in this condition.
Further, we humans also could not tell the groundtruth grouping relations by simply observing the coordinates information during $T_\mathrm{past}$ shown in the middle of \FIG{fig_threshold}.

Based on what we calculate among the ETH-UCY dataset, we design the threshold $d_\mathrm{m}$ to be 20.
Despite occasional errors, it can exclude the near ``unrelated'' agents and the off-centered ``related'' agents at the same time.

\section{Additional Analysis of Contribution Ratio}

\begin{figure}[t]
    \centering
    \includegraphics[width=1.0\linewidth]{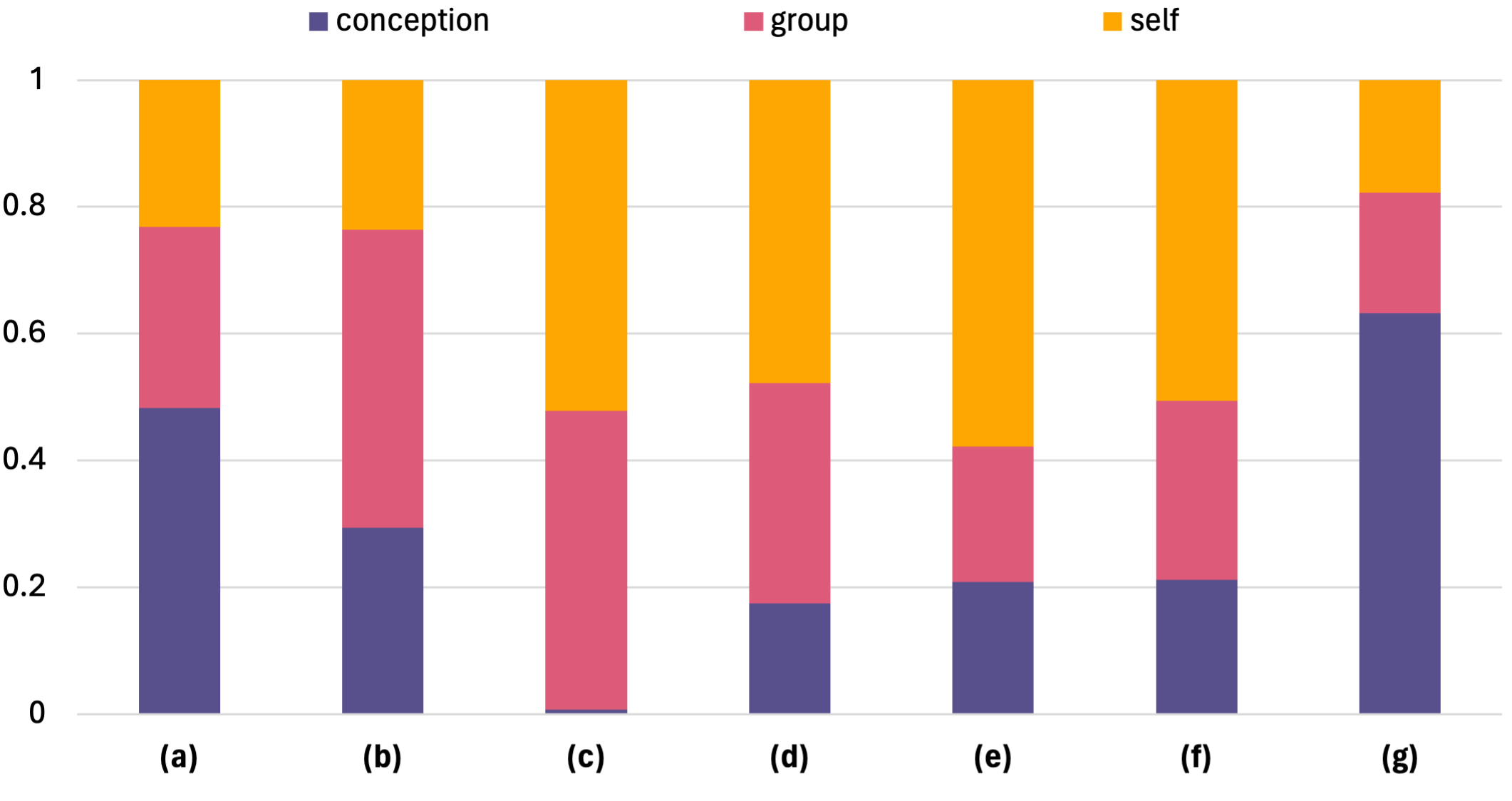}
    \caption{
        Contribution ratio of the \textcolor[RGB]{88,80,141}{\textbf{Conception}} feature, the \textcolor[RGB]{222,90,79}{\textbf{Group}} feature and the \textcolor[RGB]{255,166,00}{\textbf{Self}} feature in the form of bar charts.
        Each bar corresponds to Fig. 5 in the main manuscript.
    }
    \label{fig_contribution_ratio}
\end{figure}

The visualization results in this manuscript present the contribution ratio of self, group, and contribution in a concentric fan chart form.
By comparing the angle of each concentric fan, we can observe which feature contributes the most and which plays little role in predicting future trajectories.
Here, we further visualize this contribution ratio in a bar chart form, which can present a more accurate difference in contribution ratio.
\FIG{fig_contribution_ratio} (g) represents the largest contribution ratio in conception.
This aligns with the situation that \FIG{fig_contribution_ratio} (g) stands for the NBA dataset, where abundant interactions can be observed.
Combining the fan charts in the manuscript and the bar charts here (\FIG{fig_contribution_ratio}), we can better understand how pedestrians' decisions to make movements originate from these three features.



\end{document}


\begin{table}[htbp]
    \centering
    \footnotesize
    \begin{tabular}{|ccc|}
        \toprule
        Model & Parameter & $t$(ms) \\

        \midrule
        SocialCircle\cite{wong2023socialcircle} & 1.99M & 32 \\
        SocialCircle+\cite{wong2024socialcircle+} & 1.99M & 96 \\
        \midrule
        GPCC~\textbf{(Ours)} & 2.06M & 26 \\

        \bottomrule
    \end{tabular}
    \caption{
        Time efficiency comparisons with other methods.
        $t$ represents the average inference time tested on the ETH-UCY dataset for the models to predict 100 target agents' future trajectories.
    }
    \label{tab_time}
\end{table}